\documentclass{article}

\usepackage{microtype}
\usepackage{graphicx}
\usepackage{subcaption}
\usepackage{adjustbox}
\usepackage{booktabs} % for professional tables

\usepackage{hyperref}
\usepackage{mathtools}

\usepackage{enumitem}

\usepackage[accepted]{icml2021_arxiv}

\usepackage[usenames,dvipsnames]{xcolor}

\icmltitlerunning{Generating images with sparse representations}

\begin{document}

\twocolumn[
\icmltitle{Generating images with sparse representations}



\icmlsetsymbol{equal}{*}

\begin{icmlauthorlist}
\icmlauthor{Charlie Nash}{dm}
\icmlauthor{Jacob Menick}{dm}
\icmlauthor{Sander Dieleman}{dm}
\icmlauthor{Peter W. Battaglia}{dm}
\end{icmlauthorlist}

\icmlaffiliation{dm}{DeepMind, London, United Kingdom}

\icmlcorrespondingauthor{Charlie Nash}{charlienash@google.com}

\icmlkeywords{Machine Learning, ICML}

\vskip 0.3in
]



\printAffiliationsAndNotice{}  % leave blank if no need to mention equal contribution

\begin{abstract}
The high dimensionality of images presents architecture and sampling-efficiency challenges for likelihood-based generative models. Previous approaches such as VQ-VAE use deep autoencoders to obtain compact representations, which are more practical as inputs for likelihood-based models. We present an alternative approach, inspired by common image compression methods like JPEG, and convert images to quantized discrete cosine transform (DCT) blocks, which are represented sparsely as a sequence of DCT channel, spatial location, and DCT coefficient triples. We propose a Transformer-based autoregressive architecture, which is trained to sequentially predict the conditional distribution of the next element in such sequences, and which scales effectively to high resolution images. On a range of image datasets, we demonstrate that our approach can generate high quality, diverse images, with sample metric scores competitive with state of the art methods. We additionally show that simple modifications to our method yield effective image colorization and super-resolution models. 
\end{abstract}

\section{Introduction}
\label{sec:introduction}

\begin{figure}[h]

\begin{center}
\centerline{\includegraphics[width=\columnwidth]{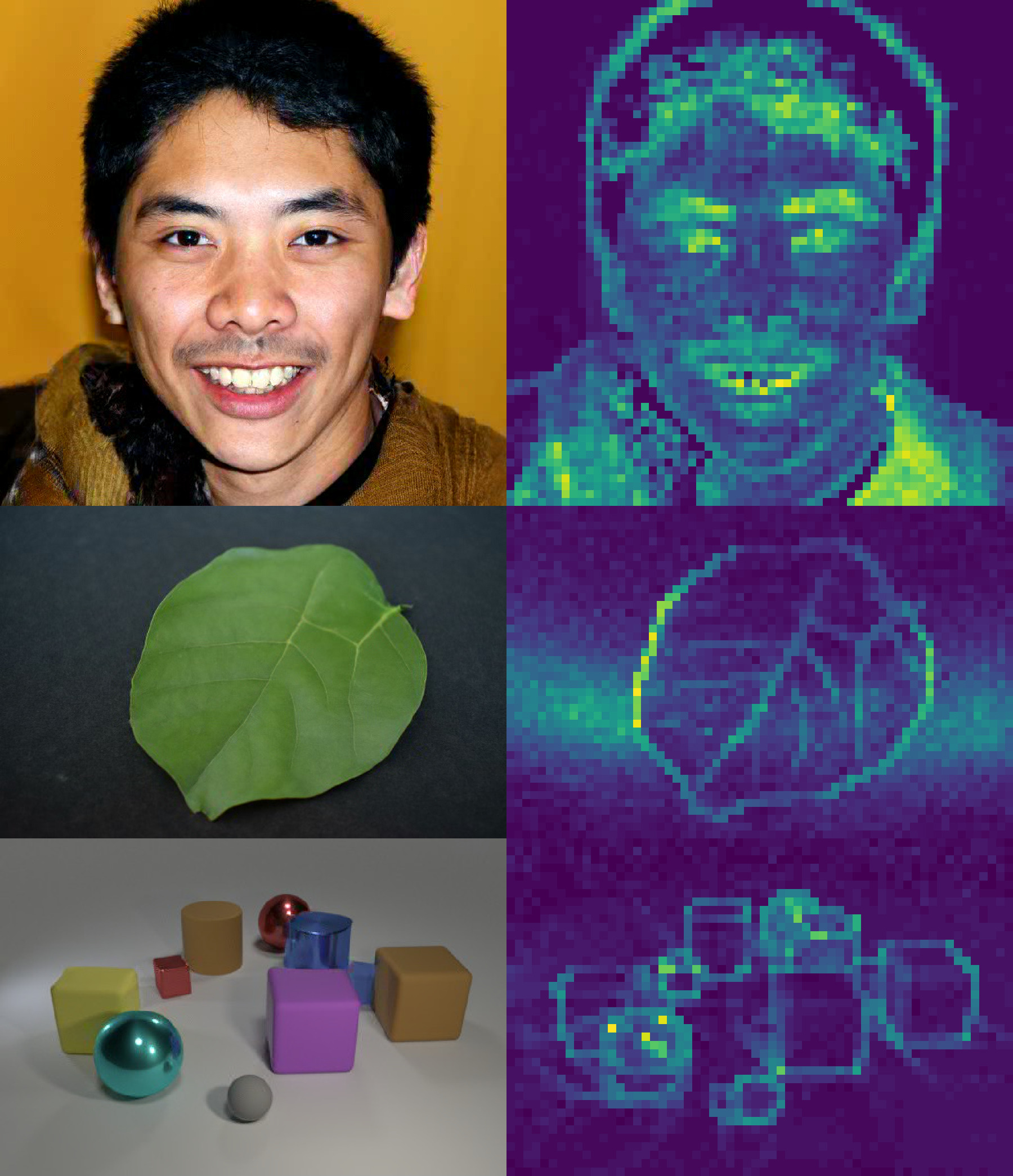}}
\vskip -0.05in
\caption{Images often have sparse structure, with salient content distributed unevenly across locations. Our model DCTransformer predicts \textit{where} to place content, as well as \textit{what} content to add. (left) Images generated by DCTransformer, and (right) associated heatmaps of image locations selected in the generation process.}
\label{fig:intro}
\end{center}
\vskip -0.325in
\end{figure}

Deep generative models of images are neural networks trained to output synthetic imagery. Current models generate sample images that are difficult for humans to distinguish from real images, and have found applications in image super-resolution~\cite{dlss}, colorization~\cite{deloldify}, and text-guided generation~\cite{dalle}. 
Such models fall broadly into three categories: generative adversarial networks (GANs,~\citealt{goodfellow2014generative}), likelihood-based models, and energy-based models. GANs use discriminator networks that are trained to distinguish samples from generator networks and real examples. Likelihood-based models, including variational autoencoders (VAEs, \citealt{kingma2014auto, rezende2014stochastic}), normalizing flows \cite{rezende2015variational}, and autoregressive models \citep{van2016pixel}, directly optimize the model log-likelihood or the evidence lower bound. Energy-based models estimate a scalar energy for each example that corresponds to an unnormalized log-probability, and can be trained with a variety of objectives~\cite{ebms}.

Likelihood-based models offer several important advantages. The training objective incentivizes learning the full data distribution, rather than only a subset of the modes (termed ``mode-dropping''), which is a common downside of GANs. Likelihood-based training tends to be more stable than adversarial alternatives, and the objective can be used to detect overfitting using a held-out set. The dramatic recent advances in modelling natural language made by GPT-3~\cite{brown2020language} demonstrate that optimizing a simple log-likelihood objective, using expressive models and large datasets, is a effective approach for modelling complex data.

Optimizing the likelihood of pixel-based images can be problematic due to their complexity and high dimensionality, however. To our knowledge, no likelihood-based model operating on raw pixels has demonstrated competitive sample quality on the ImageNet dataset~\cite{russakovsky2015image} at a resolution of 256x256 or higher. For autoregressive models, conditioning on, and sequentially sampling, the hundreds of thousands of pixels in a typical image can be prohibitive. Some likelihood-based approaches address this by reducing the dimensionality using, e.g., low-precision or quantized color spaces and images~\cite{kingma2018glow,chen2020generative,van2016pixel}. VQ-VAE \citep{oord2017neural, razavi2019generating} uses a vector-quantized autoencoder network to first perform neural lossy compression before downstream generative modelling, which reduces representation size while maintaining quality.

Beyond generative models, the dimensionality of images poses a challenge whenever data storage, transmission, and processing budgets are at a premium. Fortunately, natural images have tremendous redundancy\footnote{\citet{kersten1987predictability} estimated 4-bit grayscale pixel images have (an upper bound of) 1.42 median bits of information per pixel, which means they can generally be compressed to $\frac{1.42}{4} = 35.5\%$ their original size. They used human vision as a model, analogous to how \citet{shannon1951prediction} measured the redundancy of English by having people guess the next character in a text sequence.}, which all modern image compression methods exploit: even the images in this paper's PDF file are stored in a compressed format.

JPEG~\cite{wallace1992jpeg} and other popular lossy image compression methods use the discrete cosine transform (DCT)~\cite{ahmed1974discrete,rao1990discrete} to separate spatial frequencies of an image, and encode them with controllable resource budgets (e.g., the ``quality'' parameter). This takes advantage of the statistical structure of natural images (e.g., smooth signals with high frequency noise) and human vision (e.g., low frequencies tend to be more perceptually salient) by dropping high frequency information, to strike favorable efficiency/quality trade-offs~\cite{wang2004image}.

To capitalize on the decades of engineering behind modern compression tools, we propose a generative model over DCT representations of images, rather than pixels. We convert an image to a 3D tensor of quantized DCT coefficients, and represent them sparsely as a sequence of 3-tuples that encode the DCT channel, spatial location, and DCT coefficient for each non-zero element in the tensor. We present a novel Transformer-based autoregressive architecture~\cite{vaswani2017attention}, called ``DCTransformer'', which is trained to sequentially predict the conditional distribution over the next element in the sequence, resulting in a model that predicts both where to add content in a image, and what content to add (Figure \ref{fig:intro}). We find that the DCTransformer's sample quality is competitive with state-of-the-art GANs, but with better sample diversity. We find sparse DCT-based representations help mitigate the inference time, memory, and compute costs of traditional pixel-based autoregressive models, as well as the time-consuming training of neural lossy compression embedding functions used in models usch as VQ-VAE~\cite{oord2017neural}.

\begin{figure*}[t]
\begin{center}
\centerline{\includegraphics[width=\textwidth]{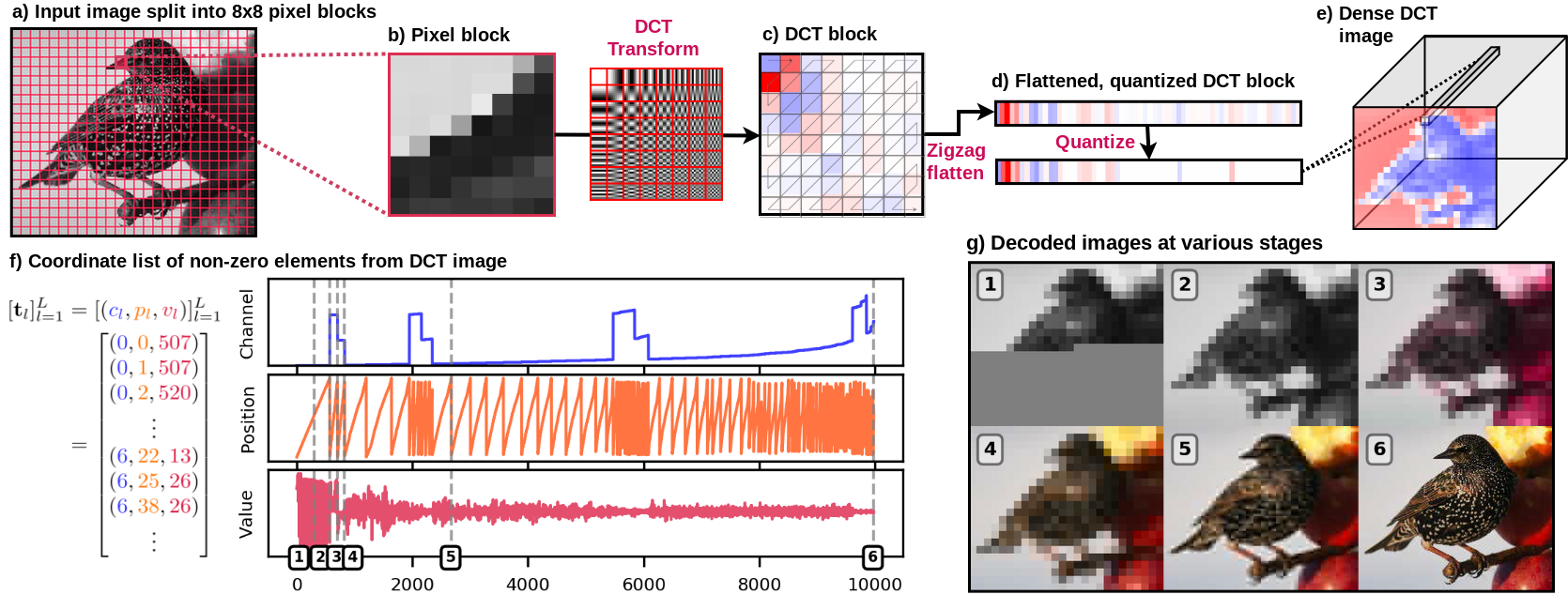}}
\caption{a) The input image is split into 8x8 blocks of pixels. b-c) The symmetric 2D DCT is used to transform each block into frequency coefficients. d) The block is flattened using the indicated zigzag ordering, and then quantized using a quality-parameterized quantization vector. e) The collection of quantized DCT vectors can be reformed into a DCT image, with 8x8=64 channels. f) The DCT image is converted to a coordinate list of non-zero channel-position-value tuples, as a sparsified representation of the image. g) Images decoded at intermediate steps indicated on the coordinate list in f). 
}
\label{fig:dct_preproc}
\end{center}
\vskip -0.3in
\end{figure*}

\section{DCT-based sparse image representations}
\label{sec:sparse_image_representations}

\subsection{Block DCT}
\label{subsec:block_DCT}
The DCT projects an image into a collection of cosine components at differing 2D frequencies. The two-dimensional DCT is typically applied to zero-centered $B \times B$ pixel blocks $\mathbf{P}$ to obtain a $B \times B$ DCT block $\mathbf{D}$:
\vspace{-.2cm}
\begin{align}
% \vskip -0.2in
    D_{uv}& = \frac{1}{4}\alpha(u)\alpha(v) \times \\
    &\sum_{x=0}^{B - 1} \sum_{y=0}^{B - 1} P_{xy} \text{cos}\left[\frac{(2x + 1)u\pi}{2B} \right]\text{cos}\left[\frac{(2y + 1)v\pi}{2B} \right]\nonumber \\
    \alpha(u) &= \begin{cases}
    \frac{1}{\sqrt(2)},& \text{if } u = 0\\
    1,              & \text{otherwise}
\end{cases}\nonumber,
\end{align}
where $x$ and $y$ represent horizontal and vertical pixel coordinates, $u$ and $v$ index the horizontal and vertical spatial frequencies, and $\alpha$ is a normalizing scale factor to enforce orthonormality. We follow the JPEG codec and split images into the YCbCr color space, containing a brightness component Y (luma) and two color components Cb and Cr (chroma). We perform 2x downsampling in both the horizontal and vertical dimensions for the two chroma channels, and then apply the DCT independently to blocks in each channel. Chroma downsampling in this way substantially reduces color information, at a minimal perceptual cost.

\subsection{Quantization}
\label{subsec:quantization}
In the JPEG codec, DCT blocks are quantized by dividing elementwise by a quantization matrix, and then rounding to the nearest integer. The quantization matrix is typically structured so that higher frequency components are squashed to a larger extent, as high frequency variation is typically harder to detect than low frequency variation. We follow the same approach, and use quality-parameterized quantization matrices for $8\times8$ blocks as defined by the \citet{ijg} in all our experiments. For block sizes other than 8, we interpolate the size 8 quality matrix. Appendix \ref{subsec:quantization_matrices} has more details.

\subsection{Sparsity}
\label{subsec:sparsity}
Sparsity is induced in DCT blocks through the quantization process, which squashes high frequency components and rounds low values to zero. The JPEG codec takes advantage of this sparsity by flattening the DCT blocks using a zigzag ordering from low to high frequency components, and applying a run-length encoding to compactly represent the strings of zeros that typically appear at the end of the DCT vectors.

We use an alternative sparse representation, where zigzag-flattened DCT vectors are reassembled in their corresponding spatial positions into a DCT image, of size $H / B \times W / B \times B^2$, where $H$ and $W$ are the height and width of the image, respectively, and $B$ is the block size (Figure \ref{fig:dct_preproc}e). The DCT image's non-zero elements are then serialized to a sparse list consisting of (DCT channel, spatial-position, value) tuples. This is repeated for the Y, Cb, and Cr image channels, with the Y channel occupying the first $B^2$ DCT bands, and the Cb and Cr channels occupying the second and third $B^2$ DCT bands, respectively. The spatial positions of the $2\times$ downsampled chroma channels are multiplied by 2 to take them into correspondence with the Y channel positions. The resulting lists are concatenated, and a stopping token is added in order to indicate the end of the variable length sequences. 

For unconditional image generation we sort from low to high frequencies, with color channels interleaved at intervals with the luminance channel. This results in a natural upsampling structure, where low frequency content is represented first, and high frequency content is progressively added, as shown in Figure~\ref{fig:dct_preproc}g. Note, this ordering is not the only option: An alternative that we discuss in Section~\ref{subsec:colorization_upsampling} places the luma data before the chroma data, resulting in an image colorization scheme. 

\begin{figure}[t]
\begin{center}
\centerline{\includegraphics[width=\columnwidth]{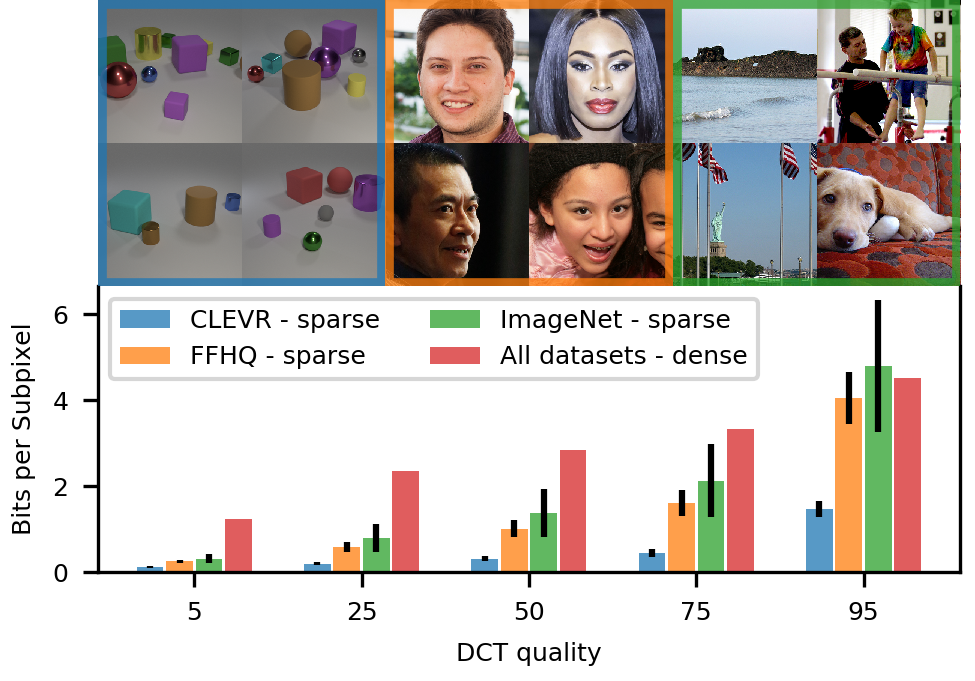}}
\vskip -0.1in
\caption{Bits per image-subpixel for dense and sparse block-DCT representations, quantized at various quality settings, reported for 1000 images from CLEVR \cite{johnson2017clevr}, FFHQ \cite{karras2019style} and ImageNet \cite{russakovsky2015image}. Dense representations use the same amount of bits in each image location, whereas sparse representations use more bits on regions of greater detail, resulting in variable length codes. For spatially sparse image datasets like CLEVR, sparse DCT representations are substantially more compact, even at higher DCT quality settings.}
\label{fig:bits}
\end{center}
\vskip -0.4in
\end{figure}

Figure \ref{fig:bits} compares the size of dense and sparse representations as a function of the DCT quality setting, and shows that for all but the highest quality setting, sparse representations are substantially smaller.

\begin{figure}[t]
% \vskip 0.2in
\begin{center}
\centerline{\includegraphics[width=\columnwidth]{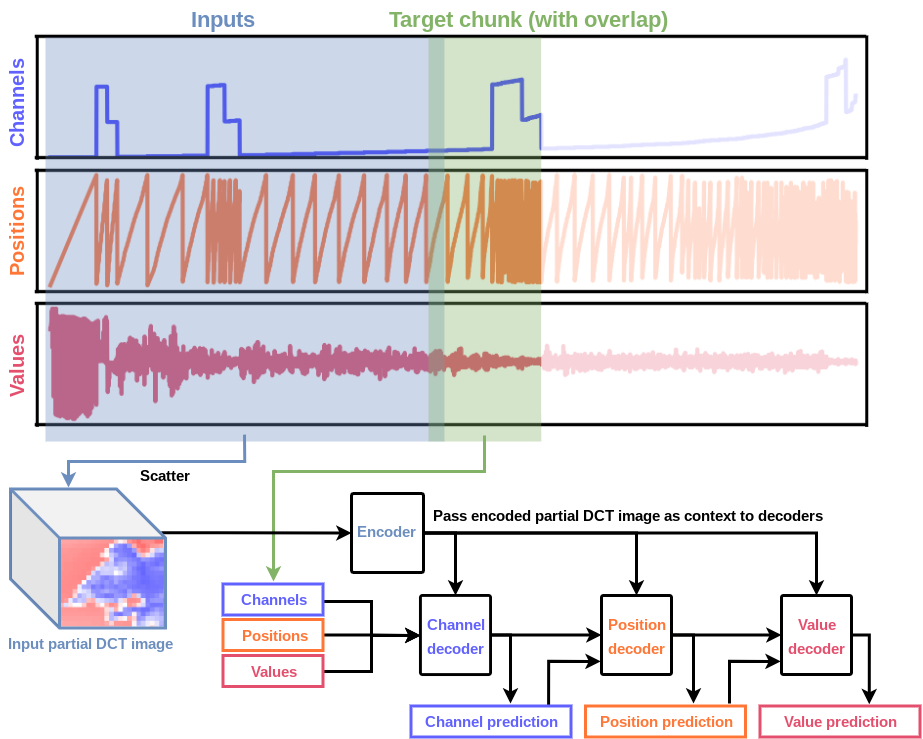}}
\caption{Chunk-based training and stacked Transformer architecture. A target chunk is selected during training, and previous elements are collected in an input slice. The input slice is gathered into a 3D DCT image Tensor. The DCT-image is flattened and embedded using a Transformer encoder, and the target slice is passed through a series of right-masked Transformer decoders, each of which performs cross-attention into the encoded DCT image, in order to predict channels, positions and values.}
\label{fig:model}
\end{center}
\end{figure}

\section{DCTransformer}
\label{sec:DCTransformer}
We model the sparse DCT sequence autoregressively, by predicting the distribution of the next element conditioned on the previous elements. Samples are generated by iteratively sampling from the predictive distributions, and decoding the resulting sparse DCT sequence to images. For a sequence consisting of channel, position, value tuples $[\mathbf{t}_l]_{l=1}^L = [(c_l, p_l, v_l)]_{l=1}^L$ the joint distribution factors as:
\begin{align}
    \prod_{l=1}^L p(c_l | \mathbf{t}_{<l} ; \theta)p(p_l | c_l, \mathbf{t}_{<l} ; \theta)p(v_l | c_l, p_l, \mathbf{t}_{<l} ; \theta), \label{eq:chain_rule}
\end{align}
where $\theta$ represents model parameters. While pixel-based autoregressive models predict values at every spatial location, DCTransformer predicts the channel of the next value, then the spatial position, and finally the quantized DCT value itself. We use categorical predictive distributions for each of the variable types, picking upper and lower bounds for the quantized DCT values.

\subsection{Chunked training for long sequences}
\label{subsec:chunked_training}
Although sparse DCT representations are more compact than raw pixels, the length of DCT sequences depends on the resolution, dataset and quality setting of the quantization matrix used. For the highest resolution datasets we consider, DCT sequences can contain upwards of 100k tuples. This presents a challenge for standard Transformer-based architectures, as the memory requirements of self attention layers scale quadratically with the sequence length. In practice, memory-issues are common in Transformers when operating on sequences with more than a few thousand elements. While memory-efficient Transformer variants \cite{child2019generating, Kitaev2020reformer, Choromanski2021rethinking, dhariwal2020jukebox, zaheer2021big, roy2020efficient} make long-sequence training more practical, care is still required to optimize model hyperparameters against available memory when training across datasets with varying sequence lengths. 

Our approach is to train on fixed-size target chunks, while conditioning on prior context in a memory and compute-efficient way (Figure ~\ref{fig:model}). During training a fixed-size target chunk is randomly selected from the DCT sequence. The sequence elements prior to the target chunk are treated as inputs, and using the inverse of the sparsification process shown in Figure \ref{fig:dct_preproc}e-f, are converted to a dense DCT image. The DCT image is optionally downsampled, then spatially flattened and embedded using a Transformer encoder. The target slice is processed with a Transformer decoder, that alternates between right-masked self-attention layers, cross attention layers that attend to the embedded DCT image, and dense layers. We use a target chunk size of 896 in all our experiments, and add an overlap of size 128 into the input chunk, so that the first predictions in the target chunk can attend directly into their preceding elements. 

The resulting architecture, shown in Figure~\ref{fig:model} uses constant memory and compute with respect to the size of the input slice, enabling training on large sequences, as well as simplifying the practical application of the model to datasets of varying resolutions. For more information about our strategy for selecting target chunks see Appendix \ref{sec:training_details}.

\subsection{Stacked channel, position, and value decoders}
\label{subsec:stacked_decoders}
In order to model the three distinct variable types in DCT coordinate lists, our architecture must yield autoregressive predictions for each variable, and should condition on all the available context. One possible approach is to flatten the tuples, and to pass the resulting values through a single Transformer decoder, projecting the resulting states to logits as appropriate for each variable. To increase the amount of content modelled in a particular target sequence, we instead opt to use three distinct Transformer decoders, one for each of the channel, position and value predictions. The three decoders are stacked on top of one another, yielding sequence lengths three times smaller than flattened sequences.

\newcommand{\aspectheight}{1.6in}
\newcommand{\squareheight}{1.4in}
\newcommand{\skipsize}{2.025cm}
\begin{figure*}[h!]
     \begin{center}
     \begin{tabular}{  cccc  }
    %  \hline
      \adjustbox{valign=c}{\includegraphics[height=\aspectheight]{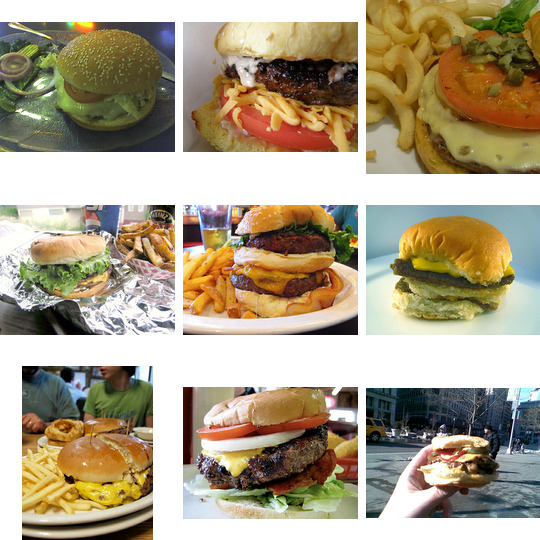}}
      & 
      \adjustbox{valign=c}{\includegraphics[height=\aspectheight]{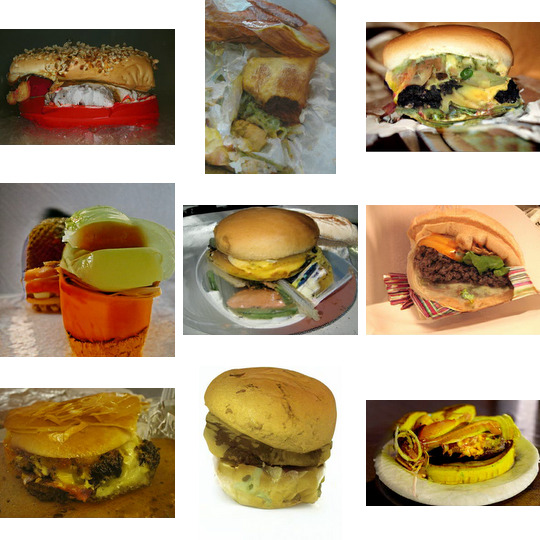}}
      &
      \adjustbox{valign=c}{\includegraphics[height=\squareheight]{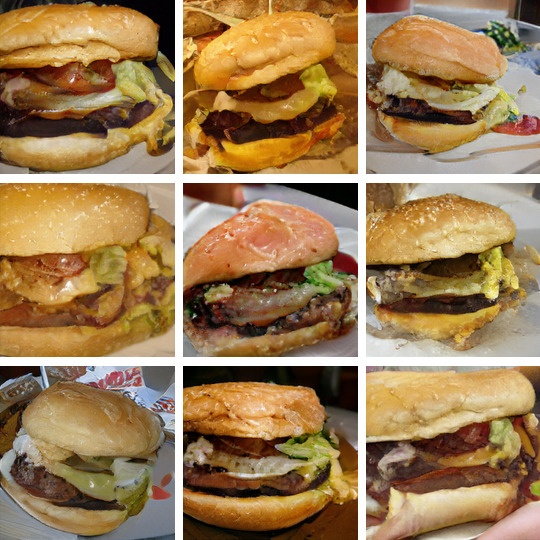}}
      &
      \adjustbox{valign=c}{\includegraphics[height=\squareheight]{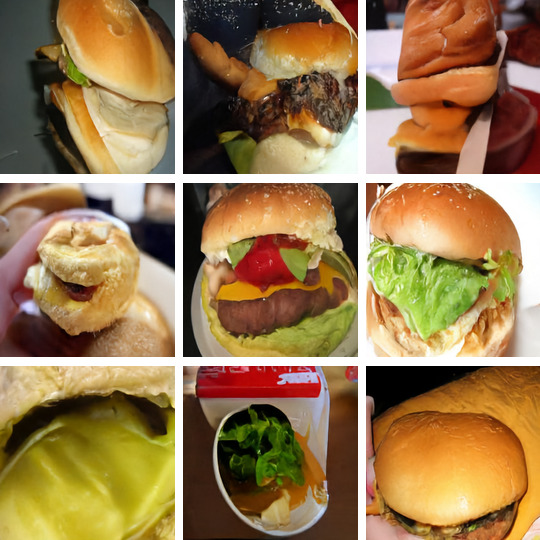}}
      \\[\skipsize]
      \adjustbox{valign=c}{\includegraphics[height=\aspectheight]{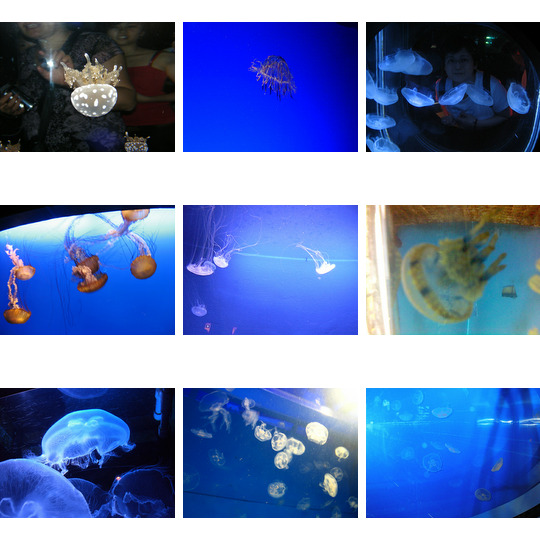}}
      & 
      \adjustbox{valign=c}{\includegraphics[height=\aspectheight]{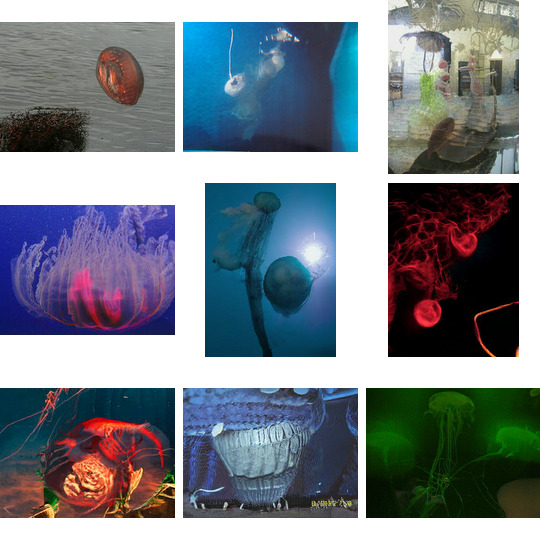}}
      &
      \adjustbox{valign=c}{\includegraphics[height=\squareheight]{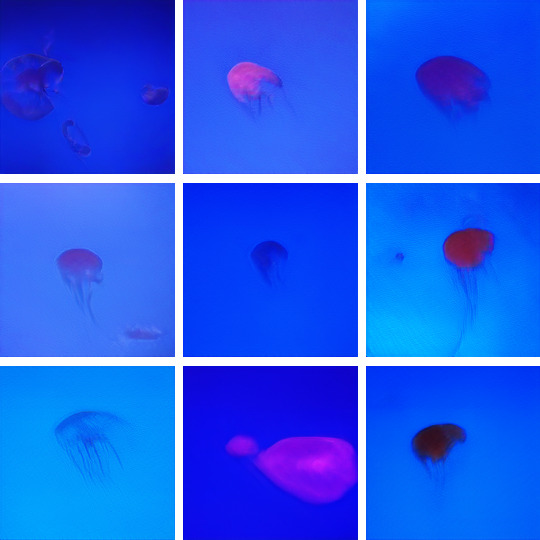}}
      &
      \adjustbox{valign=c}{\includegraphics[height=\squareheight]{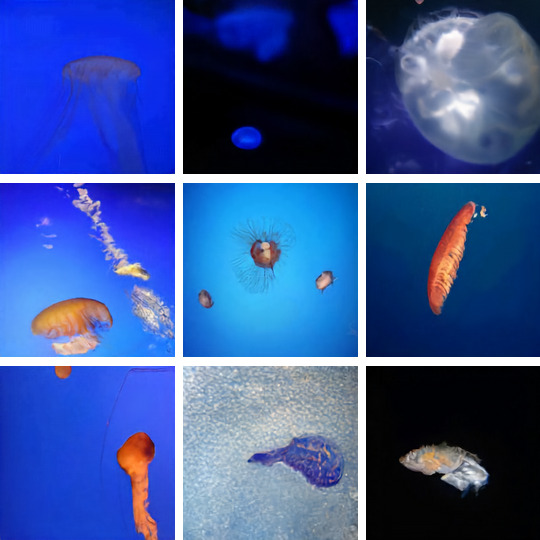}}
      \\[\skipsize]
      \adjustbox{valign=c}{\includegraphics[height=\aspectheight]{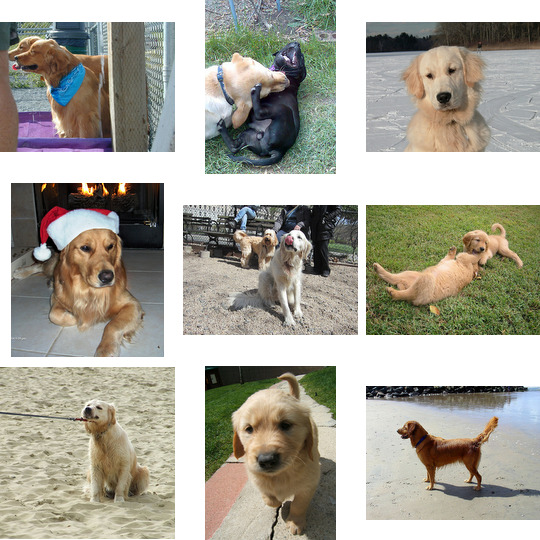}}
      & 
      \adjustbox{valign=c}{\includegraphics[height=\aspectheight]{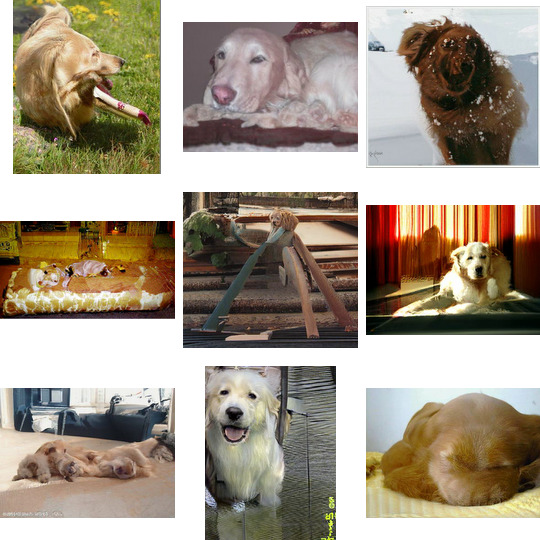}}
      &
      \adjustbox{valign=c}{\includegraphics[height=\squareheight]{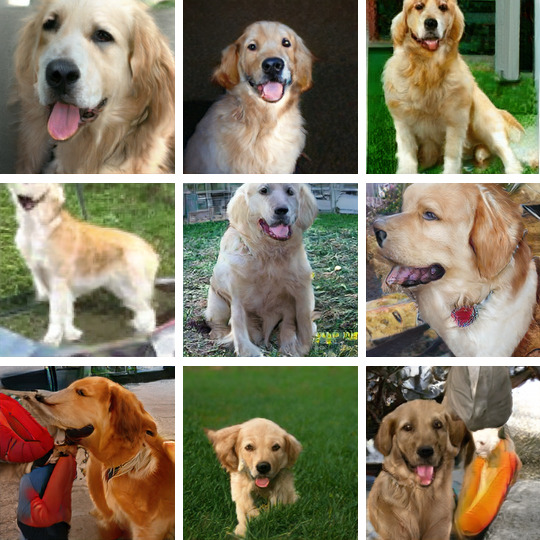}}
      &
      \adjustbox{valign=c}{\includegraphics[height=\squareheight]{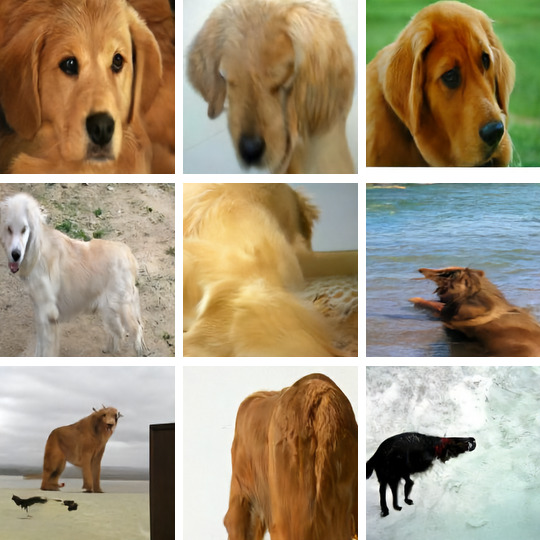}}
      \\ 
      Data & DCTransformer & BigGAN Deep & VQ-VAE2 \\
      \end{tabular}
      \caption{Comparison between ImageNet images and model samples for cheeseburger (label-933), jellyfish (label-107) and golden retriever (label-207) classes. DCTransformer produces variable aspect ratio samples with long-side resolution 384. BigGAN and VQ-VAE are trained and sample at a fixed 256x256 resolution corresponding to a resized long-side crop of the input images. BigGAN samples use truncation 1.0 to yield maximum diversity.}
      \label{fig:imagenet_comparison}
      \end{center}
      \vskip -0.15in
\end{figure*}

Let $\textbf{D}$ be the input DCT image associated with a given input slice, then DCT image embeddings are obtained by passing the flattened DCT image through a Transformer encoder:
\begin{align}
    \textbf{E}_{\text{input}} = \text{encode}\left(\textbf{D}_{\text{flat}} \right).
\end{align}
Predictive distributions for the channels are obtained by passing the encoded DCT image along with summed-channel $\mathbf{C}$, image position $\mathbf{P}$, value $\mathbf{V}$, and chunk-position $\mathbf{P}_{\text{chunk}}$ embeddings into the channel decoder:
\begin{align}
    \mathbf{E}_{\text{channel}} &= \mathbf{C}_{1:S-1} + \mathbf{P}_{1:S-1} + \mathbf{V}_{1:S-1} + \mathbf{P}_{\text{chunk}}\\
    \mathbf{H}_{\text{channel}} &= \text{decode}_{\text{channel}}\left( \mathbf{E}_{\text{channel}} \ ; \ \mathbf{E}_{\text{input}}\right).
\end{align}
We distinguish here between inputs that are processed with masked self attention (left), and those that are processed with cross-attention (right). The final hidden state $\mathbf{H}_{\text{channel}}$ is passed along with updated channel embeddings to the position decoder:
\begin{align}
    \mathbf{E}_{\text{position}} &= \mathbf{H}_{\text{channel}} + \mathbf{C}_{2:S} \\
    \mathbf{H}_{\text{position}} &= \text{decode}_{\text{position}}\left( \mathbf{E}_{\text{position}} \ ; \ \mathbf{E}_{\text{input}}\right)
\end{align}
When predicting a particular DCT value, we know its associated channel, and spatial position (Equation \ref{eq:chain_rule}). We use this information to gather values from the embedded DCT inputs that are in the same spatial position as the target DCT value. This allows the value decoder to access all of the prior DCT values in the spatial position where it is making a prediction. The gathered DCT embeddings are added to the prior hidden state $\mathbf{H}_{\text{position}}$ and passed to the value decoder:
\begin{align}
    \mathbf{E}_{\text{value}} &= \mathbf{H}_{\text{position}} + \text{gather}\left(\mathbf{E}_{\text{inputs}}, \mathbf{P}_{2:S} \right) \\
    \mathbf{H}_{\text{value}} &= \text{decode}_{\text{value}}\left( \mathbf{E}_{\text{value}} \ ; \ \mathbf{E}_{\text{input}}\right)
\end{align}
The resulting hidden states $\mathbf{H}_{\text{channel}}$, $\mathbf{H}_{\text{position}}$ and $\mathbf{H}_{\text{value}}$ are each passed through linear layers to obtain the logits of their associated predictive distributions.

\begin{figure*}[ht!]
\centering
\begin{subfigure}[t]{\textwidth}
\centering
\includegraphics[width=\textwidth]{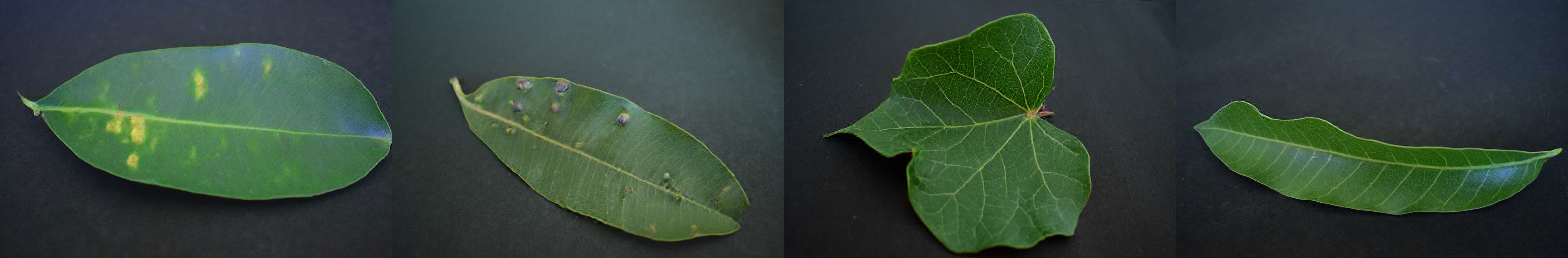}
\caption{\textbf{Plant leaves} \cite{plant-leaves} Long-side resolution 2048, block-size 32 and DCT quality 50. Available at {\footnotesize{\url{https://www.tensorflow.org/datasets/catalog/plant_leaves}}}}
\end{subfigure}
\begin{subfigure}[t]{\textwidth}
\centering
\includegraphics[width=\textwidth]{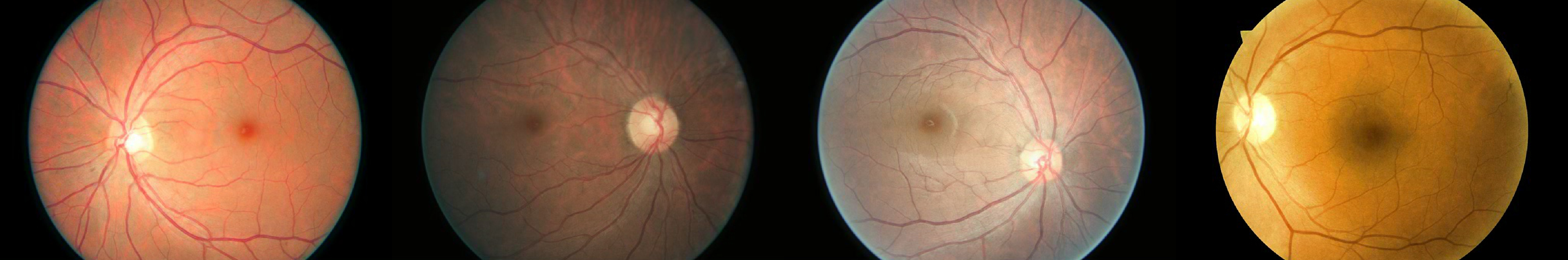}
\caption{\textbf{Diabetic retinopathy} \cite{kaggle-diabetic-retinopathy} Long-side resolution 1024, block-size 16 and DCT quality 75. Available at {\footnotesize{\url{https://www.tensorflow.org/datasets/catalog/diabetic_retinopathy_detection}}}}
\end{subfigure}
\begin{subfigure}[t]{\textwidth}
\centering
\includegraphics[width=\textwidth]{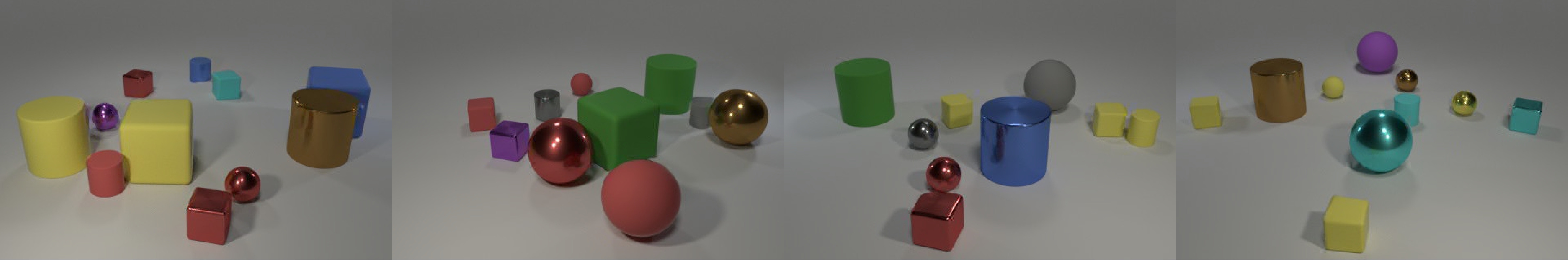}
\caption{\textbf{CLEVR} \cite{johnson2017clevr} Long-side resolution 480, block-size 8 and DCT quality 90. Available at {\footnotesize{\url{https://www.tensorflow.org/datasets/catalog/clevr}}}}
\end{subfigure}
\begin{subfigure}[t]{\textwidth}
\centering
\includegraphics[width=\textwidth]{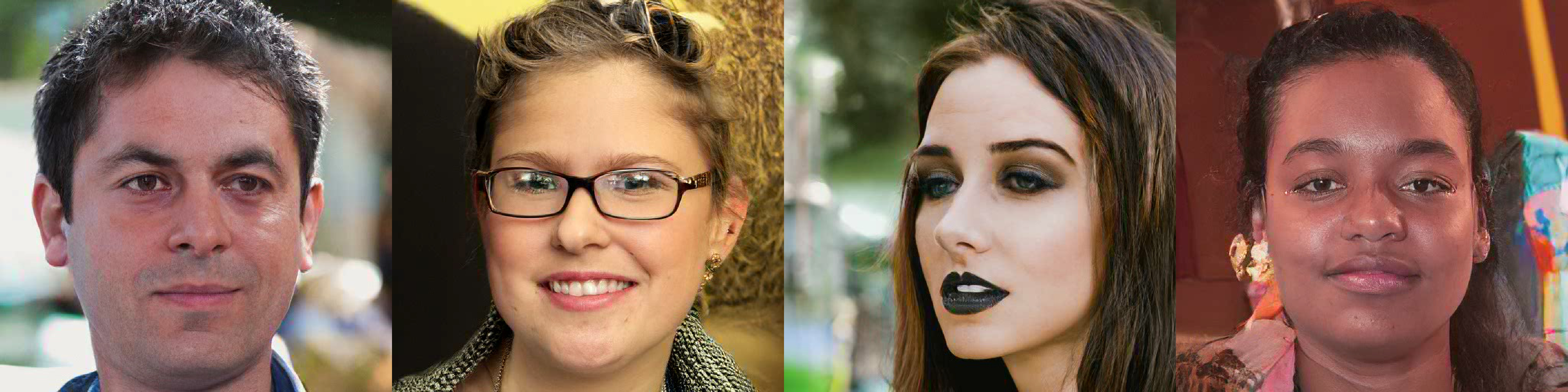}
\caption{\textbf{FFHQ} \cite{karras2019style} Long-side resolution 1024, block-size 16 and DCT quality 35 Available at {\footnotesize{\url{https://github.com/NVlabs/ffhq-dataset}}}}
\end{subfigure}
\caption{Selected image samples on a range of varied image datasets. Each set of 4 images is selected from a sampled batch of 50 to showcase the image quality achievable by DCTransformer. 
\label{fig:images_diverse}
}
% \vskip -0.1in
\end{figure*}

\begin{table}[ht!]
\centering
\begin{tabular}{@{}l@{\hspace{0.5\tabcolsep}}cccc@{}}
\toprule
Model               & Precision            & Recall               & FID &  sFID                  \\ 
\\
\multicolumn{4}{@{} l}{\textbf{LSUN Bedrooms}} & \multicolumn{1}{l}{} \\ \midrule
StyleGAN            & \textbf{0.55}        & 0.48                 & \textbf{2.45}   &  6.68   \\     \\
ProGAN              & 0.43                 & 0.40                  & 8.35        &  9.46       \\
DCTransformer              & 0.44                 & \textbf{0.56}        & 6.40    &    \textbf{6.66}         \\
                    & \multicolumn{1}{l}{} & \multicolumn{1}{l}{} & \multicolumn{1}{l}{} & \multicolumn{1}{l}{} \\
\multicolumn{4}{@{} l}{\textbf{LSUN Towers}} & \multicolumn{1}{l}{} \\ \midrule
ProGAN              & 0.51                 & 0.33                 & 10.24   &       10.02         \\
DCTransformer              & \textbf{0.54}        & \textbf{0.54}        & \textbf{8.78}  &     \textbf{7.70}   \\
                    & \multicolumn{1}{l}{} & \multicolumn{1}{l}{} & \multicolumn{1}{l}{}  & \multicolumn{1}{l}{} \\
\multicolumn{4}{@{} l}{\textbf{LSUN Churches}} & \multicolumn{1}{l}{} \\ \midrule
StyleGAN2           & \textbf{0.60}         & 0.43                 & \textbf{4.01}  &    \textbf{9.28}    \\
ProGAN              & 0.61                  & 0.38                 & 6.42       &     10.47     \\
DCTransformer              & 0.60              & \textbf{0.48}        & 7.56         &  10.71      \\
                    & \multicolumn{1}{l}{} & \multicolumn{1}{l}{} & \multicolumn{1}{l}{}  & \multicolumn{1}{l}{} \\
\textbf{FFHQ} & \multicolumn{1}{l}{} & \multicolumn{1}{l}{} & \multicolumn{1}{l}{}  & \multicolumn{1}{l}{} \\ \midrule
StyleGAN              & \textbf{0.71}               & \textbf{0.41}                & \textbf{4.39}      &    11.57     \\
DCTransformer              & 0.51        & 0.40        & 13.06   &  \textbf{9.44}          \\
                    & \multicolumn{1}{l}{} & \multicolumn{1}{l}{} & \multicolumn{1}{l}{}  & \multicolumn{1}{l}{} \\
\multicolumn{4}{@{} l}{\textbf{ImageNet (class conditional)}} & \multicolumn{1}{l}{} \\ \midrule
BigGAN-deep           & \textbf{0.78}     & 0.35                & \textbf{6.59}   &   \textbf{6.66}     \\
VQ-VAE2            & 0.36        & 0.57                & 31.11   &  17.38      \\
DCTransformer              & 0.36        & \textbf{0.67}        & 36.51     &   8.24               \\
\bottomrule
\end{tabular}
\caption{Precision, recall and FID metrics comparison. sFID is equivalent to FID but uses intermediate spatial features in the inception network rather than the spatially-pooled features used in standard FID.}
\label{tbl:results}
\vskip -0.3in
\end{table}

\subsection{Sampling}
\label{subsec:sampling}
As with training, sampling operates in a chunked fashion. Given a sequence of observed values from a DCT coordinate list, the model constructs an input dense DCT block, and proceeds to autoregressively sample a fixed size continuation. Once the chunk is complete its values are added to the input DCT image, and the process is repeated until a maximum number of chunks have been produced, or the stopping token is sampled. As with training, the sampling process requires constant computation and memory for each block. This contrasts with standard Transformers where these factors expand with the sequence length. 

\section{Experiments}
\label{sec:experiments}
% We train DCTransformers on a range of datasets. In each case
Quantitative comparison of generative image models is challenging, as not all model classes define normalized likelihoods (e.g., GANs), and even for models that do, the underlying data representation may not be the same, and so likelihoods are not directly comparable. In VQ-VAE and related models for example, likelihood scores are meaningful only in relation to the latent codes associated with a particular neural encoder. We therefore use the sample-based Frechet inception distance (FID)~\cite{heusel2017gans} and precision and recall~\cite{kynk2019improved} metrics, which enable comparison between any model that produces samples. We report FID using both the standard \texttt{pool\_3} inception features, and the first 7 channels from the intermediate \texttt{mixed\_6/conv} feature maps, which we refer to sFID \cite{szegedy2016rethinking}. \texttt{pool\_3} features compress spatial information to a large extent, making them less sensitive to spatial variability. We include the intermediate \texttt{mixed\_6/conv} features to provide a picture of the spatial distributional similarity between models. We use the first 7 channels in order to obtain a feature space of size $17 \times 17 \times 17 = 2023$, which is comparable size to the size 2048 \texttt{pool\_3} feature vector. We report Precision and Recall scores based on \texttt{pool\_3} features. For all FID scores we follow \citet{brock2019large} and compare 50k model samples to features computed on the entire training set.  

\begin{table}[t]
\centering
\begin{tabular}{@{}l@{\hspace{0.5\tabcolsep}}cccc@{}}
\toprule
Model               & Precision            & Recall               & FID &  sFID                  \\ 
& \multicolumn{1}{l}{} & \multicolumn{1}{l}{} & \multicolumn{1}{l}{}  & \multicolumn{1}{l}{} \\
\multicolumn{4}{@{} l}{\textbf{ImageNet (class conditional 8x upsampling)}}   & \multicolumn{1}{l}{} \\ \midrule
Val.~set (low res.~)           & 0.02        & 0.09                & 163.13   &  314.27      \\
Val.~set            & \textbf{0.69}        & 0.59                & \textbf{5.72}   &  14.06      \\
DCTransformer              & 0.61        & \textbf{0.61}        & 9.98     &   \textbf{11.06}               \\
& \multicolumn{1}{l}{} & \multicolumn{1}{l}{} & \multicolumn{1}{l}{}  & \multicolumn{1}{l}{} \\
\multicolumn{4}{@{} l}{\textbf{OpenImagesV4 colorization}}   & \multicolumn{1}{l}{} \\ \midrule
Val.~set (greyscale)          & 0.61        & 0.28                & 34.19   &  38.09     \\
Val.~set              & \textbf{0.75}        & 0.40        & 26.95     &   27.29               \\
DCTransformer           & 0.72        & \textbf{0.41}                & \textbf{22.52}   &  \textbf{22.83}      \\
\bottomrule
\end{tabular}
\caption{Precision, recall, and FID metrics for upsampling and colorization models. }
\label{tbl:other_results}
\end{table}

\subsection{Image generation benchmarks}
\label{subsec:image_gen_benchmarks}

Table~\ref{tbl:results} shows FID, precision, recall scores on LSUN datasets \cite{yu15lsun}, Flickr faces HQ (FFHQ, \citet{karras2019style}) as well as class-conditional ImageNet \cite{russakovsky2015image}. On the LSUN datasets we compare to ProGAN \cite{karra2018pro} and StyleGAN \cite{karras2019style, karras2020style} baselines, using publically available repositories of  samples from pre-trained models. For LSUN towers samples from a pre-trained StyleGAN model are not publically available. On Imagenet we compare to BigGAN \cite{brock2019large}, and VQ-VAE2 \cite{razavi2019generating}, using sample repositories sent to us by the VQ-VAE2 authors. 

We find that overall the picture is mixed, with GANs achieving the strongest precision and FID scores, but DCTransformer typically achieving the best recall scores, reflecting the likelihood training metric that emphasises data coverage. The FID gap is closed to a large extent when using the spatial sFID, with DCTransformer achieving the best scores on three of the five datasets studied. This is consistent with our observation that DCTransformer samples tend to be diverse, and to exhibit realistic textures and structures, but are somewhat less reliably coherent than GAN samples. Figure \ref{fig:imagenet_comparison} shows uncurated samples from three ImageNet classes, and Figures \ref{fig:ffhq_comparison} and \ref{fig:lsun_comparison} in appendix \ref{subsec:additional_samples} show uncurated samples on FFHQ and LSUN datasets respectively. For model hyperparameters and training details see appendix \ref{sec:training_details}. 

\begin{figure*}[h!]
\begin{center}
\centerline{\includegraphics[width=\textwidth]{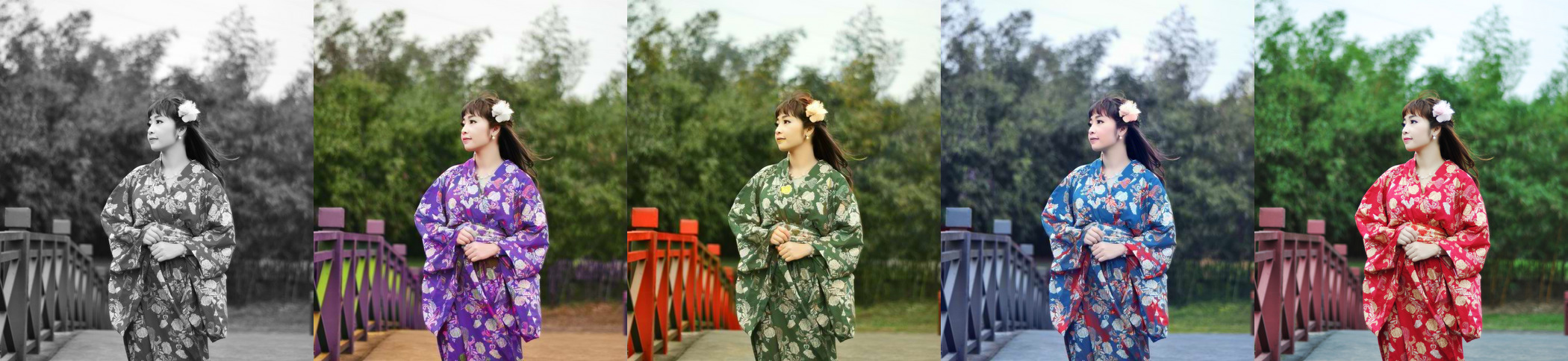}}
\vskip -0.1in
\caption{Image colorization. (left) grayscale image, (middle) three samples generated by DCTransformer, (right) original image.}
\label{fig:colorization}
\end{center}
\vskip -0.2in
\end{figure*}

\begin{figure*}[h!]
\begin{center}
\centerline{\includegraphics[width=\textwidth]{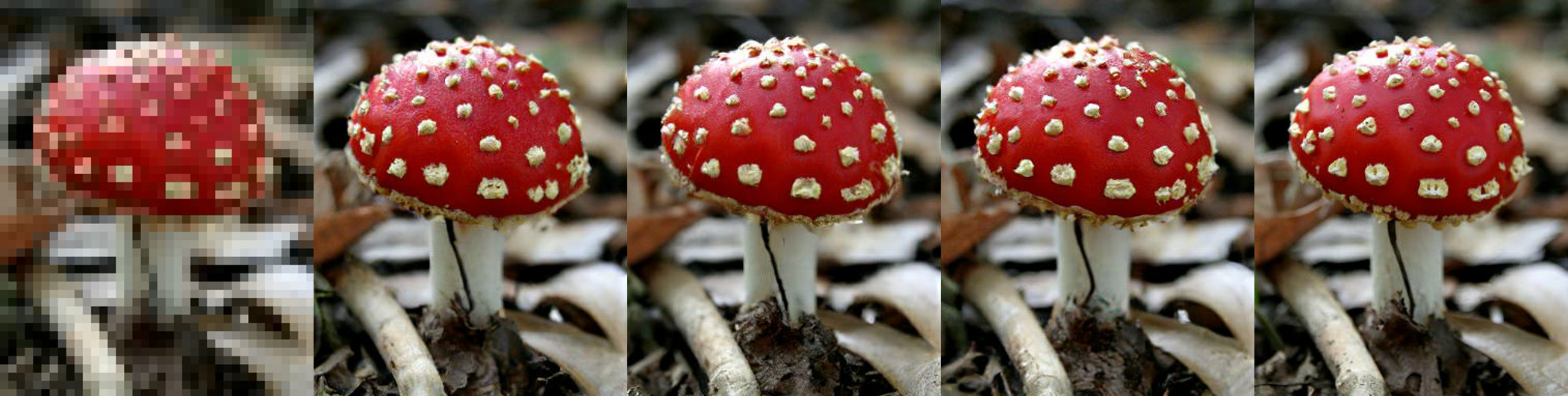}}
\vskip -0.1in
\caption{8x image upsampling. (left) downsampled image, (middle) three samples generated by DCTransformer, (right) original image.}
\label{fig:upsampling}
\end{center}
\vskip -0.2in

\end{figure*}

\subsection{Image generation on diverse datasets}
\label{subsec:image_gen_diverse}
Due to the resolution independence of our chosen image representation and architecture, we can apply the same model to images that vary in resolution. Unlike GANs, our models are very stable in training, and enable us to detect overfitting by evaluating log-likelihood scores. As such DCTransformer is straightfoward to apply to new image datasets. To illustrate the flexibility of our model, we train on a diverse set of image datasets, ranging from synthetic 3D objects to medical images of the retina. Figure~\ref{fig:images_diverse} shows curated samples selected from a batch of 50 examples. We note the high quality results obtained on the plant leaves dataset at long-side resolution 2048. To our knowledge these are the first high-quality unconditional generations at this resolution in the literature using neural networks. 

\subsection{Image colorization and upsampling}
\label{subsec:colorization_upsampling}
The ordering we use for image generation naturally results in a model that upsamples from low frequency DCT components to high frequency~(see~Section~\ref{subsec:sparsity}). We investigate the effectiveness of DCTransformer as a class-condtitional upsampling model by providing the first DCT channel for the luminance and chrominance components as a context, and sampling sequence continuations using a model trained on ImageNet for class-conditional generation. The first DCT component represents the average intensity in a pixel block, and so this procedure is equivalent to upsampling by a factor of the block size. Figure \ref{fig:upsampling} shows some example results, and Figure \ref{fig:upsampling_multiples} in appendix \ref{subsec:additional_samples} shows additional uncurated results. Table \ref{tbl:other_results} shows sample metrics for the upsampling model, where we compare DCTransformer-upsampled validation set images to the original validation set. The table shows that with the exception of the precision metric, DCTransformer-upsampled images obtain scores similar or better than to the validation set images.  

In our standard DCT sequence ordering, color channels and luminance channels are interleaved, which means color information is distributed throughout the sequence. If we instead place the color channels at the end of the sequence and condition on all the luminance information, we can train DCTransformer as a colorization model. We train a model on OpenImages V4 \cite{kuznetsova2018open}, and unlike the upsampling model, we train only on target slices from the color channels at the end of the sequence. Figure \ref{fig:colorization} shows some example results, and Figure \ref{fig:colorization_multiples} in Appendix \ref{subsec:additional_samples} shows additional uncurated results. Table~\ref{tbl:other_results} shows sample metric scores for DCTransformer, compared to ground truth, as well as the greyscale components of the ground truth images. The sample metrics are very close to the validation set scores, suggesting a close distributional match to the training distribution. 

\section{Related Work}
\label{sec:related_work}

Our method builds on prior autoregressive generative models of images. Perhaps the most similar work is VQ-VAE2 \cite{razavi2019generating}, which is also an autoregressive model of an image representation that has undergone lossy compression. In the case of VQ-VAE2, the image representation has fewer dimensions than the raw pixels but always a fixed size, whereas our sparse representation has a dynamic size depending on the content. VQ-VAE2 uses a hierarchy of latent codes, each modelled with a PixelSnail-style network with self-attention components. For the low-to-high frequency ordering we use in most experiments, a similar coarse-to-fine representation is obtained.

Other work has also found it advantageous to model a representation of the image which has lost information deemed less perceptually less relevant, e.g. representing RGB values with lower bit-depth \cite{kingma2018glow, menick18spn}. The Subscale Pixel Network (SPN) also performs autoregressive spatial upsampling, and makes use of Transformers \cite{menick18spn}. Similar to Section \ref{subsec:chunked_training}, SPN also subsamples a target sequence during training and uses a separate encoder network to condition on past context in order to obtain constant memory/compute costs with respect to image size. 

\section{Discussion}
\label{subsec:discussion}
We proposed using sparse DCT-based image representations for generative modelling, and introduce a Transformer-based autoregressive architecture for modelling sparse image sequences which overcomes challenges in modelling long sequences with a chunked training regime. Our DCTransformer achieves strong performance on sample quality and diversity benchmarks, and easily supports super-resolution upsampling, as well as colorization tasks. We believe there is much to be gained from exploring the data representations used in classical data compression methods as a basis for neural generative modelling, and plan to explore related representations in audio and video domains.

There are still some challenges: For complex and high resolution datasets, good results require large models and substantial computational resources (Appendix \ref{sec:training_details}). This contrasts with GANs in particular, which achieve high quality results with less computational resources. However, we believe the computational disparity is primarily due to the challenging likelihood maximization objective, which strongly incentivizes full coverage of the data distribution, and enables us to produce highly diverse outputs. 

\bibliography{main}
\bibliographystyle{icml2021}

\clearpage
\appendix

\section{Quantization matrices}
\label{subsec:quantization_matrices}
Following the \citet{ijg} standard we use the following quality parameterized quantization matrix $\mathbf{Q}$ to quantize DCT pixel blocks. For quality $q$ in $[1,100]$ the matrix is defined as:
\begin{align*}
\mathbf{T}_{\text{luma}} &= \begin{bmatrix}
16  &  11  &  10  &  16  &  24  &  40  &  51  &  61 \\
12  &  12  &  14  &  19  &  26  &  58  &  60  &  55 \\
14  &  13  &  16  &  24  &  40  &  57  &  69  &  56 \\
14  &  17  &  22  &  29  &  51  &  87  &  80  &  62 \\
18  &  22  &  37  &  56  &  68  & 109  & 103  &  77 \\
24  &  35  &  55  &  64  &  81  & 104  & 113  &  92 \\
49  &  64  &  78  &  87  & 103  & 121  & 120  & 101 \\
72  &  92  &  95  &  98  & 112  & 100  & 103  &  99 \\
\end{bmatrix} \\ 
s(q) &= \begin{cases}
    5000 / q, & \text{if } q < 50\\
    200 - 2q,              & \text{otherwise}
\end{cases} \\
\mathbf{Q}_{\text{luma}}(q) &= \text{floor}\left(\frac{s(q)\mathbf{T}_{\text{luma}} + 50}{100}\right)
\end{align*}

For the chrominance components we replace $\mathbf{T}_{\text{luma}}$ with an alternative base matrix $\mathbf{T}_{\text{chroma}}$ which provides stronger quantization:

\begin{align*}
\mathbf{T}_{\text{chroma}} = \begin{bmatrix}
17 & 18&	24&	47&	99&	99&	99&	99\\
18 &	21&	26&	66&	99&	99&	99&	99\\
24 &	26&	56&	99&	99&	99&	99&	99\\
47&	66&	99&	99&	99&	99&	99&	99\\
99&	99&	99&	99&	99&	99&	99&	99\\
99&	99&	99&	99&	99&	99&	99&	99\\
99&	99&	99&	99&	99&	99&	99&	99\\
99&	99&	99&	99&	99&	99&	99&	99\\
\end{bmatrix}.
\end{align*}
When using block sizes other than 8, we use nearest neighbour interpolation to resize the base matrices $\mathbf{T}_{\text{luma}}$ and $\mathbf{T}_{\text{chroma}}$ to the target size.

\section{Architecture details}
\label{sec:architecture_details}
DCTransformer consists of a Transformer encoder that processes partial DCT images, and three stacked Transformer decoders that process slices from the DCT co-ordinate list (Section \ref{subsec:stacked_decoders}). We use a number of modifications to the original architecture that we found to improve stability, training speed, and memory consumption:

\paragraph{Layer norm placement} Following \citet{child2019generating} and \citet{parisotto2020stabilizing} we use Transformer blocks with layer norm placed inside the residual path, rather than applying layer norm 

\paragraph{ReZero} We use ReZero \cite{bachlechner2020rezero, de2020batch}, multiplying each residual connection with a zero-initialized scalar value, which is optimized jointly with the model parameters. In our experiments we found this to improve training speed and stability to a small degree. The combination of ReZero and our chosen layer norm placement results in residual connections of the following form:
\begin{align}
    \mathbf{H}_l = \mathbf{H}_{l-1} + \alpha_l f_l(\text{layernorm}(\mathbf{H_{l-1}})),
\end{align}
where $\mathbf{H}_l$ is a sequence of activations at layer $l$, and $f$ is the residual function.

\paragraph{PAR Transformer} \citet{mandava2020pay} showed through Neural architecture search that the default alternation of fully connected and self attention layers in Transformer blocks is sub-optimal with respect to performance-speed trade offs. Based on the results of the search they proposed the PAR Transformer, which applies a series of fully connected layers after each self-attention layer, resulting in improved inference speed, and memory savings. We use a PAR Transformer style architecture in DCTransformer encoder and decoders, and while we didn't experiment rigorously with the ratio of fully-connected to self-attention layers, we found that architectures using 2-4 fully connected layers per self-attention layer helped to boost the parameter count at a given memory budget. 

Table \ref{tbl:hyperparams} details the architecture configurations used for our main experiments. 

\begin{table*}[h]
\centering
{\small
\begin{tabular}{@{}lllll@{}}
\toprule
                              & LSUN (all)                            & FFHQ                            & ImageNet & OpenImagesV4                    \\ \midrule
Image resolution              & 384                             & 1024                             & 384             &      640          \\
DCT block size                & 8                               & 16                               & 8                 &       8       \\
DCT quality                   & 75                              & 35                              & 75                  &   50         \\
DCT clip value & 1200 & 3200 & 1200 & 1200 \\
Target chunk size & 896 & 896 & 896 & 896 \\
Target chunk overlap & 128 & 128 & 128 & 128 \\

Hidden units                  & 896                            & 896                            & 1152              &      896        \\
Self-attention heads          & 14                              & 14                              & 14                     &    18     \\
Layer spec (encoder)          & [(1,2)] * 4                     & [(1, 2)] * 4                    & [(1,2)] * 4          &   [(1,2)] * 4          \\
Layer spec (channel decoder)  & [(1, 2)] * 3 + [(1, 4)]         & [(1, 2)] * 3 + [(1, 4)]         & [(1, 2)] * 3 + [(1, 4)]     &  [(1, 2)] * 3 + [(1, 4)]   \\
Layer spec (position decoder) & [(1, 2)] * 3 + [(1, 4)]         & [(1, 2)] * 3 + [(1, 4)]         & [(1, 2)] * 3 + [(1, 4)]     &  [(1, 2)] * 3 + [(1, 4)]   \\
Layer spec (value decoder)    & [(1, 2)] * 5 + [(1, 7)]         & [(1, 2)] * 6 + [(1, 7)]         & [(1, 2)] * 6 + [(1, 7)]     &   [(1, 2)] * 6 + [(1, 7)] \\
DCT image downsampling kernel & kernel size 4, stride 2         & kernel size 6, stride 3         & kernel size 8, stride 4    &   kernel size 6, stride 3  \\
Batch size                  & 512                            & 448                            & 512                    &   512       \\
Dropout rate                  & 0.1                            & 0.1                            & 0.01                    &   0.01       \\
Learning Rate Start           & 5e-4                            & 5e-4                            & 5e-4                    &     5e-4    \\
Tokens processed              & 300e9                          & 250e9                          & 1000e9                  &    1000e9      \\
Parameters                    & 448e6                              & 473e6                              & 738e6  &  533e6  \\
TPUv3 cores                   & 64                             & 64                             & 128            & 64 \\
\bottomrule                    
\end{tabular}
% \vspace{1cm}
\par\bigskip
\begin{tabular}{@{}llll@{}}
\toprule
                              & Plant Leaves                            & Retinopathy                            & CLEVR                        \\ \midrule
Image resolution              & 2048                             & 1024                             & 480                             \\
DCT block size                & 32                               & 16                               & 8                               \\
DCT quality                   & 50                              & 75                              & 90                              \\
DCT clip value & 4000 & 3200 & 1200 \\
Target chunk size & 896 & 896 & 896 \\
Target chunk overlap & 128 & 128 & 128 \\
Hidden units                  & 768                            & 768                            & 896                            \\
% Fully connected dim           & 4608                            & 4608                            & 4608                            \\
Self-attention heads          & 12                              & 12                              & 14                              \\
Layer spec (encoder)          & [(1,2)] * 4                     & [(1, 2)] * 4                    & [(1,2)] * 4                     \\
Layer spec (channel decoder)  & [(1, 2)] * 3 + [(1, 3)]         & [(1, 2)] * 3 + [(1, 3)]         & [(1, 2)] * 3 + [(1, 4)]         \\
Layer spec (position decoder) & [(1, 2)] * 3 + [(1, 4)]         & [(1, 2)] * 3 + [(1, 4)]         & [(1, 2)] * 3 + [(1, 4)]         \\
Layer spec (value decoder)    & [(1, 2), (1, 2), (1, 3), (1, 3), (1, 7)]       & [(1, 2), (1, 2), (1, 3), (1, 3), (1, 7)]         & [(1, 2)] * 6 + [(1, 7)]         \\
DCT image downsampling kernel & kernel size 4, stride 2         & kernel size 4, stride 2         & kernel size 6, stride 3         \\
Batch size                  & 256                            & 512                            & 128              \\
Dropout rate                  & 0.4                            & 0.1                            & 0.5                            \\
Learning Rate Start           & 5e-4                            & 5e-4                            & 5e-4                            \\
Tokens processed              & 100e9                          & 200e9                          & 50e9                          \\
Parameters                    & 325e6                              & 318e6                             & 483e6 \\
TPUv3 cores                   & 64                             & 64                             & 32        \\\bottomrule                    
\end{tabular}
}
\caption{Model and training hyperparameters. The layer spec for the Transformer encoder and decoders is a list of tuples, where each tuple describes the number of self-attention layers, followed by the number of fully-connected layers in a Transformer block. For example [(1, 2)] * 3 + [(1, 4)] expands to [(1,2),(1,2),(1,2),(1,4)], and corresponds to four Transformer blocks, where the first three blocks consist of a single self-attention layer, followed by two fully-connected layers. The final block has one self-attention layer followed by four fully-connected layers.}
\label{tbl:hyperparams}
\end{table*}

\section{Training details}
\label{sec:training_details}
\paragraph{Optimization} We train our models using Adam Optimizer \cite{kingma2014adam}  and train for a fixed number of tokens, where the number of tokens processed in a batch is the number of elements in the target chunk that we apply a loss to. We use a linear warmup of 1000 steps up to a maximum learning rate, and use a cosine decay over the course of training. We train all models using Google Cloud TPUv3 \cite{tpu}, and detail the number of cores used during training in Table \ref{tbl:hyperparams}.

\begin{figure*}[h!]
\begin{center}
\centerline{\includegraphics[width=\textwidth]{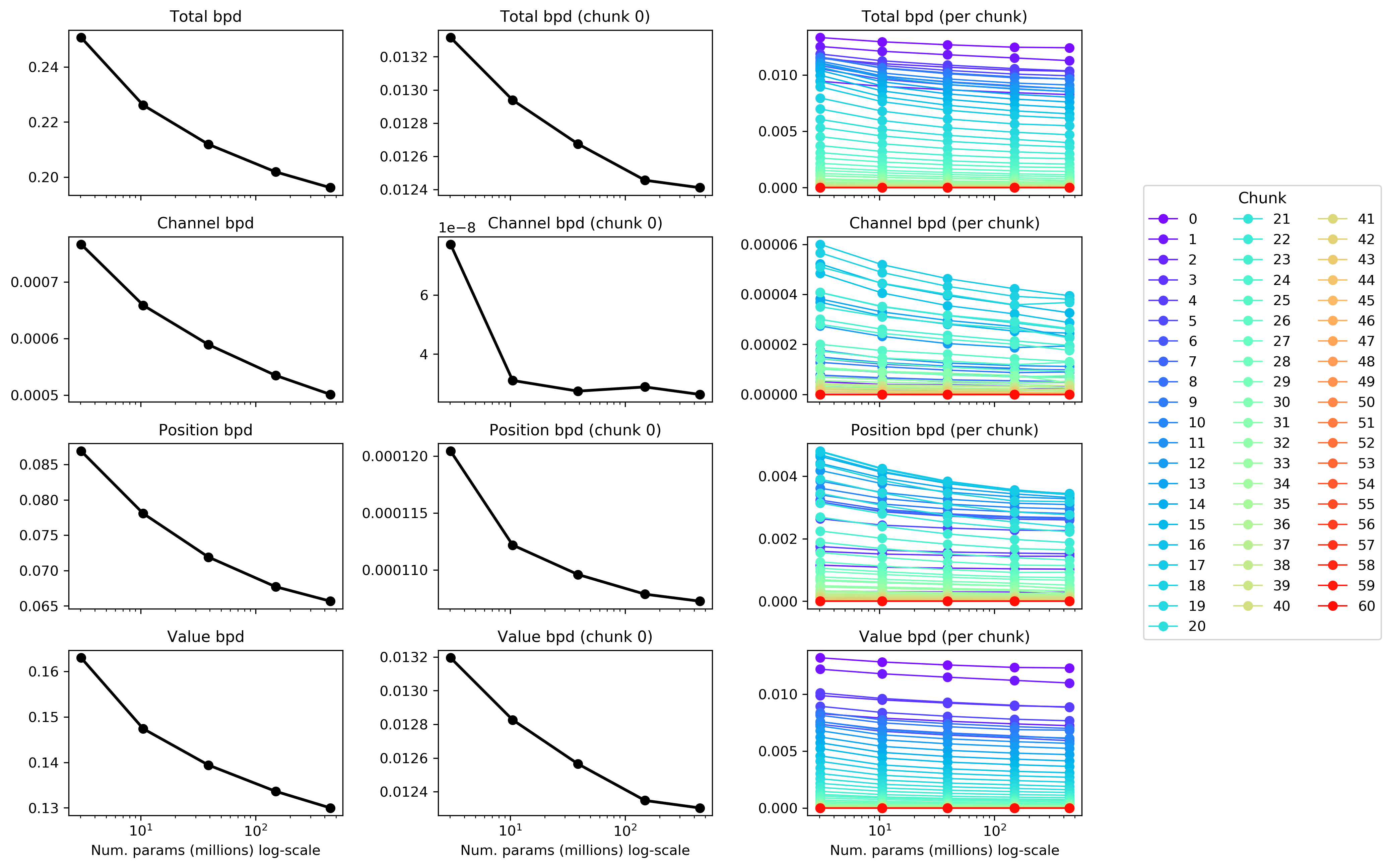}}
\caption{Model performance, reported in bits per image subpixel (bpd) as a function of model size for DCTransformer trained on LSUN bedrooms. We report the total bpd, as well as the contributions from the channel, position and value distributions. We additionally report bpd broken down by sequence chunk, where each chunk corresponds to 896 sequence elements. The model size is reported in terms of the number of total parameters, ranging from roughly 3 million to 448 million.}
\label{fig:scaling}
\end{center}
\vskip -0.85cm
\end{figure*}

\paragraph{Chunk selection policy} As described in Section \ref{subsec:chunked_training}, we train DCTransformer on input and target chunks selected from larger sequences. If the target chunk is sampled uniformly from the sequence representation of an image, the sample gradient is unbiased because it is a Monte Carlo estimate of the gradient computed on the full sequence. As discussed in Section \ref{sec:sparse_image_representations}, lossy image compression codecs such as JPEG preferentially discard high frequency data when allocating limited storage capacity. It stands to reason that we may benefit from making an analogous decision when allocating limited model capacity. We found it advantageous for sample quality to bias the selection of target chunks toward the beginning of the sequence, which contains more low frequency information (see Figure \ref{fig:dct_preproc}).

We select chunks using the following process: A sequence of length $L$ is split into chunks of size $C$, with the final chunk containing $L \text{ mod } C$ elements. The probability of selecting a chunk with start position $l$ decays to a lower limit $p_{\text{min}}$, with probability proportional to a polynomial in $l$. By default we decay the probability of selecting a chunk beginning at position $l$ proportional to $l^{-3}$ down to a minimum of $p_{\text{min}} = 0.1$, a hyperparameter choice we fixed early in model development and found no need to adjust. 

\paragraph{Sequence length bias adjustment} Chunk-based training introduces an issue in unconditional generative modelling: It biases the model towards chunks from shorter sequences. Consider the first chunk, that occurs at the very start of the sequence. For long sequences, this chunk will be selected relatively infrequently compared to short sequences. The model will therefore assign greater probability to initial chunks from short sequences, than initial chunks from long sequences.

We counter the bias by randomly filtering out sequences with a probability that is inversely proportional to the sequence length. We pick a maximum filtering sequence length $L_{\text{max}}$ and filter our sequences with probability:
\begin{align}
    p_{\text{filter}}\left([\mathbf{t}_l]_{l=1}^L\right) = \text{maximum}\left(\frac{L}{L_{\text{max}}}, 1\right).
\end{align}

\section{Model scaling properties}
\label{subsec:scaling}
\citet{kaplan2020scaling} show that for Transformer language models, performance improves reliably subject to constraints on model size, compute and dataset size. In particular, they show that if compute and data are not bottlenecks, then test-loss performance improves log-linearly with model size. Follow-up work \citet{henighan2020scaling} has shown that this phenomenon applies more generally beyond just language data. We investigate the extent to which DCTransformer scales as a function of model size by training models of varying sizes on LSUN bedrooms. Figure \ref{fig:scaling} shows the results broken down by channel, position and value prediction contributions, and chunk position. We find that the total bits-per-dimension (bpd) roughly matches the expected log-linear fit, with the exception of the smallest model, which performs worse than the expected trend. Another exception is chunk 0, where the performance improvement is less than expected for the largest model. 

We also find that the value predictions account for the largest portion of the total bpd, followed by positions, and finally channels, which accounts for a very small portion of the total bpd. For the total bpd, the contribution per chunk decreases with the chunk position. This is likely because the total bpd is dominated by the value bpd, and we expect the value bpd to decrease as we transition from lightly quantized low frequency components, to more heavily compressed high frequency components.

\section{Additional samples}
\label{subsec:additional_samples}
\begin{figure*}[t]
\begin{center}
\centerline{\includegraphics[width=\textwidth]{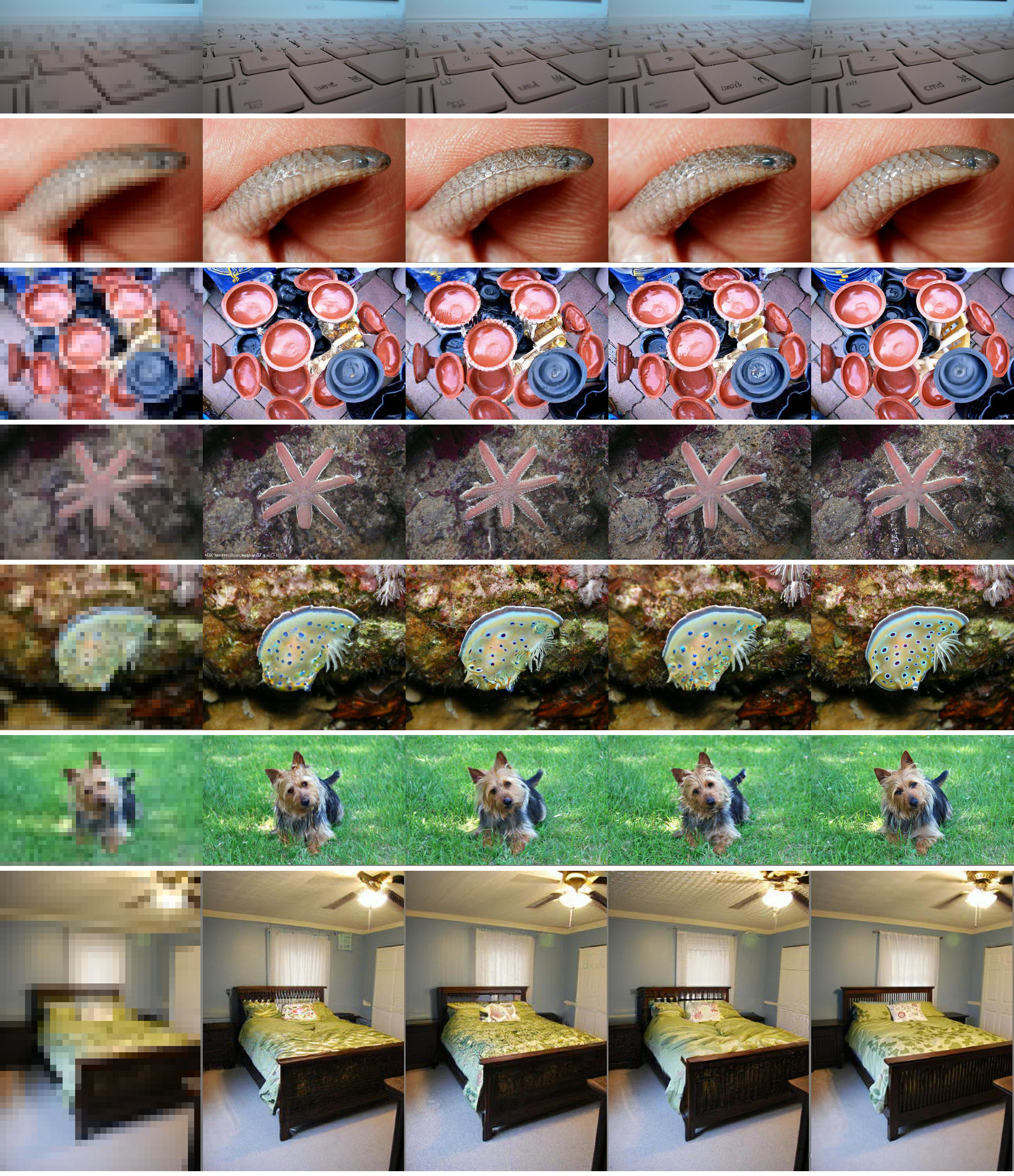}}
\caption{Uncurated 8x image upsampling results on ImageNet validation set. (left) input downsampled image, (middle) three samples generated by DCTransformer, (right) original image.}
\label{fig:upsampling_multiples}
\end{center}
\end{figure*}

\begin{figure*}[t]
\begin{center}
\centerline{\includegraphics[width=\textwidth]{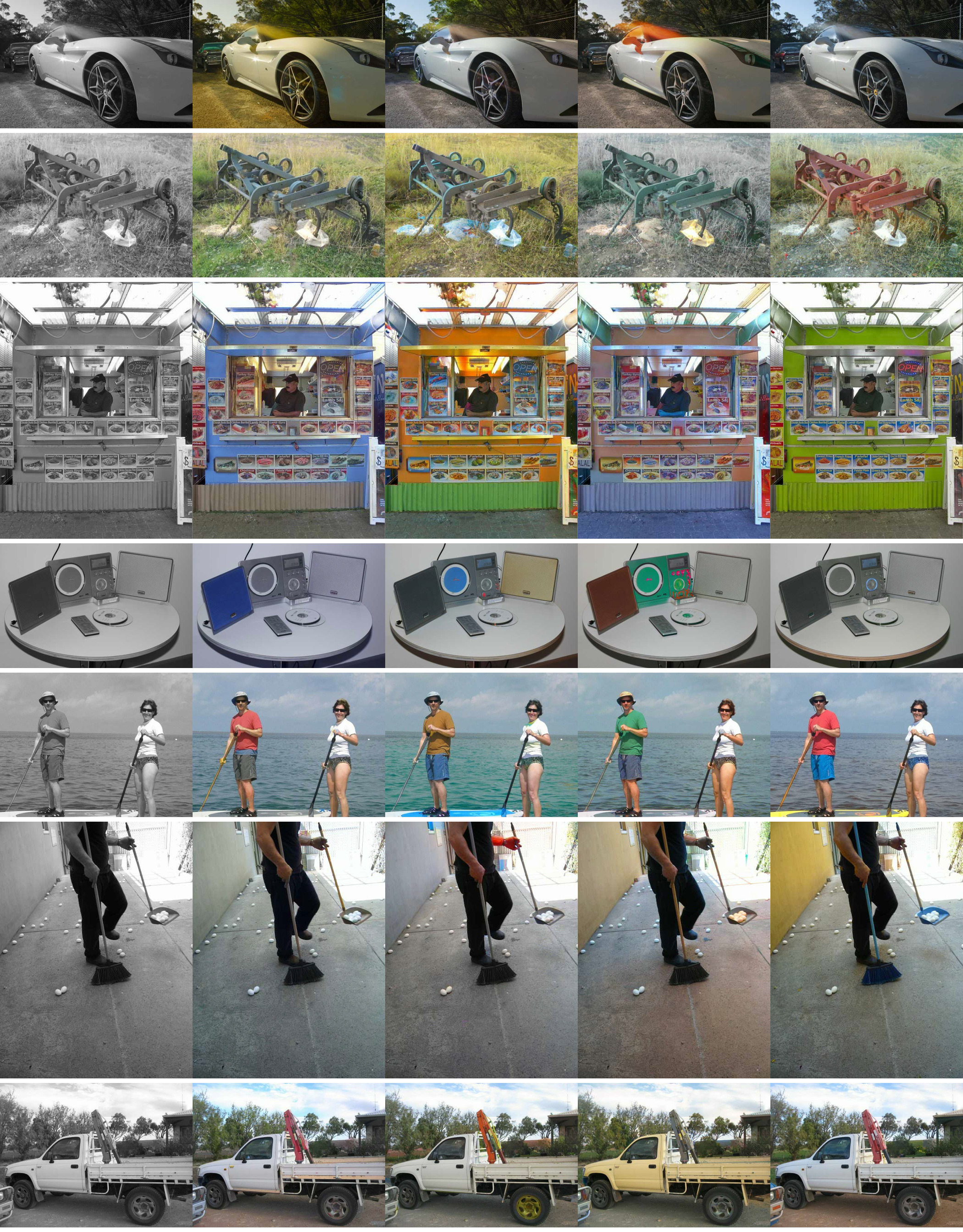}}
\caption{Uncurated image colorization results on OpenImagesV4 validation set. (left) input grayscale image, (middle) three samples generated by DCTransformer, (right) original image.}
\label{fig:colorization_multiples}
\end{center}
\end{figure*}

Figures \ref{fig:ffhq_comparison} and \ref{fig:lsun_comparison} compare uncurated samples from DCTransformer and baselines to a random selection of real data on the FFHQ and LSUN datasets respectively. Figures \ref{fig:upsampling_multiples} and \ref{fig:colorization_multiples} show uncurated upsampling and colorization results on ImageNet and OpenImagesV4 respectively.

\renewcommand{\squareheight}{2.15in}
\begin{figure*}[h]
     \begin{center}
     \begin{tabular}{  ccc  }
    %  \hline
      \adjustbox{valign=c}{\includegraphics[height=\squareheight]{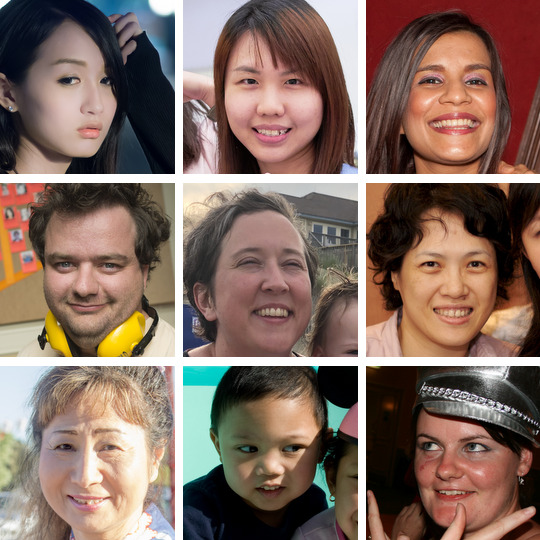}}
      & 
      \adjustbox{valign=c}{\includegraphics[height=\squareheight]{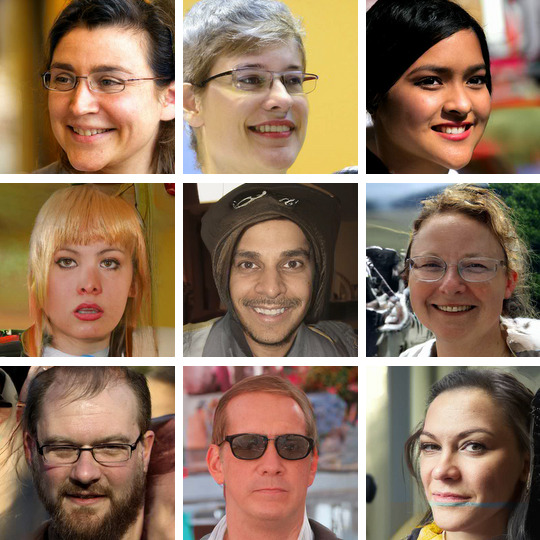}}
      &
      \adjustbox{valign=c}{\includegraphics[height=\squareheight]{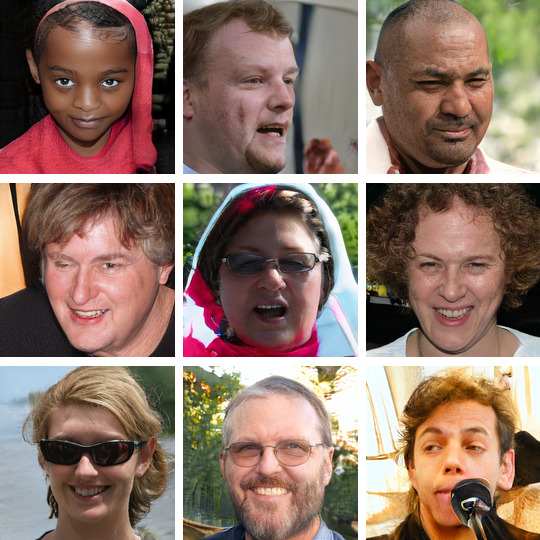}}
      
      \\  %\hline
      Data & DCTransformer & StyleGAN \\
      \end{tabular}
      \caption{Comparison between FFHQ images and uncurated model samples, all at 1024x1024 resolution.}
      \label{fig:ffhq_comparison}
      \end{center}
\end{figure*}

\renewcommand{\aspectheight}{1.6in}
\renewcommand{\squareheight}{1.4in}
\renewcommand{\skipsize}{2.025cm}
\begin{figure*}[t]
     \begin{center}
     \begin{tabular}{  cccc  }
    %  \hline
      \adjustbox{valign=c}{\includegraphics[height=\aspectheight]{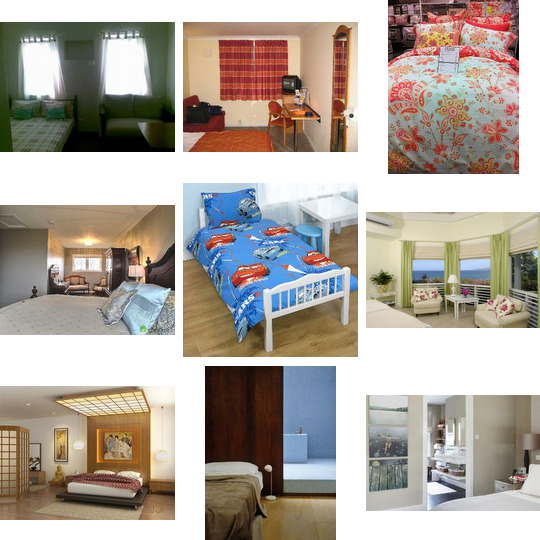}}
      & 
      \adjustbox{valign=c}{\includegraphics[height=\aspectheight]{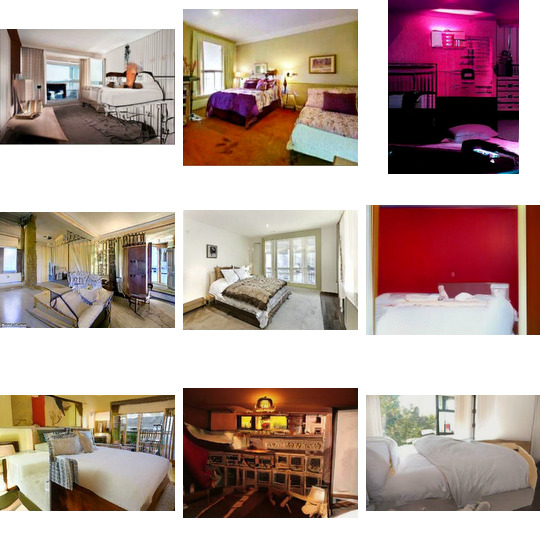}}
      &
      \adjustbox{valign=c}{\includegraphics[height=\squareheight]{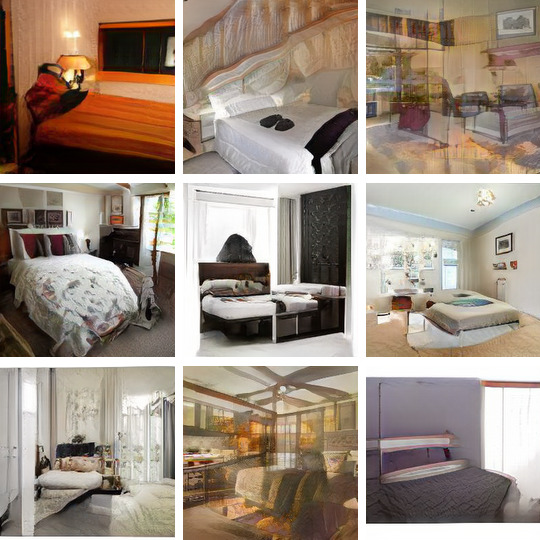}}
      &
      \adjustbox{valign=c}{\includegraphics[height=\squareheight]{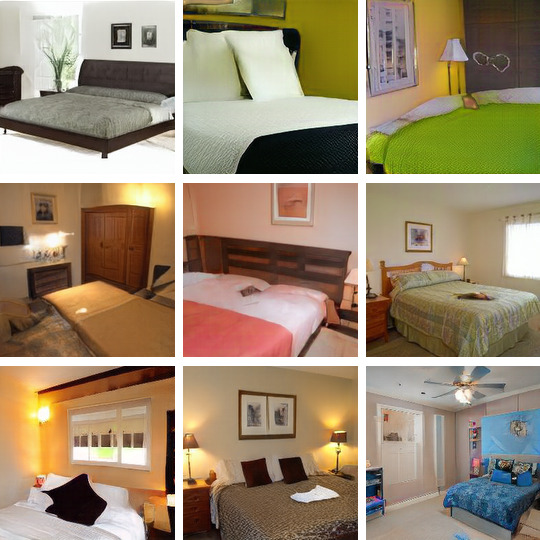}}
      \\[\skipsize]
      \adjustbox{valign=c}{\includegraphics[height=\aspectheight]{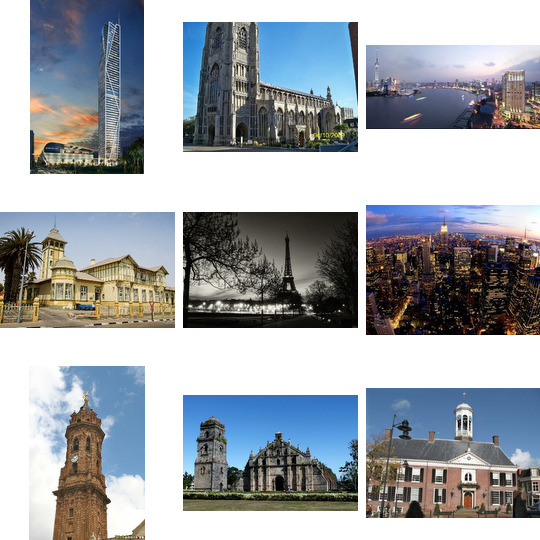}}
      & 
      \adjustbox{valign=c}{\includegraphics[height=\aspectheight]{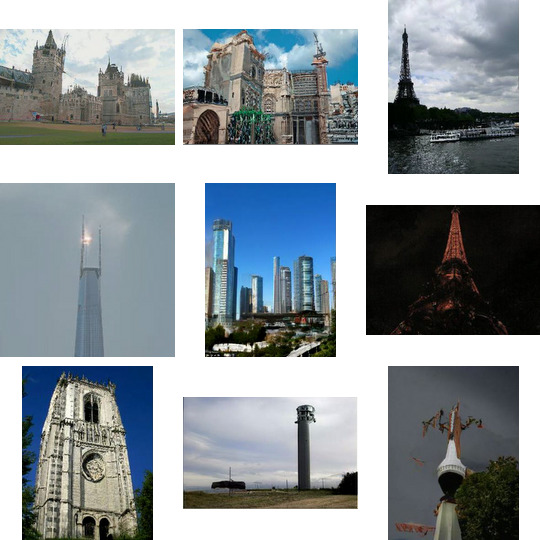}}
      &
      \adjustbox{valign=c}{\includegraphics[height=\squareheight]{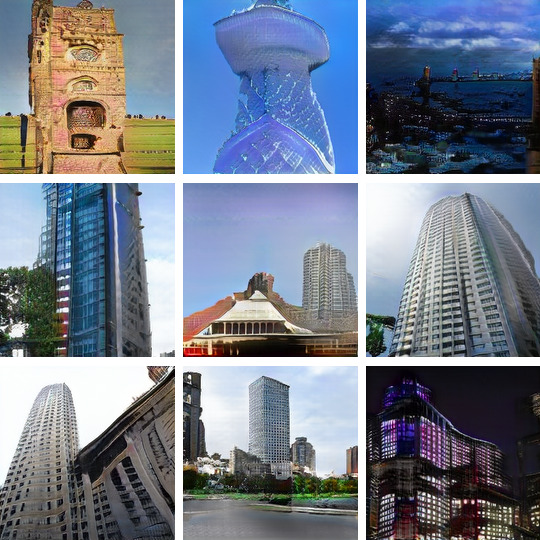}}
      &
      
      \\[\skipsize]
      \adjustbox{valign=c}{\includegraphics[height=\aspectheight]{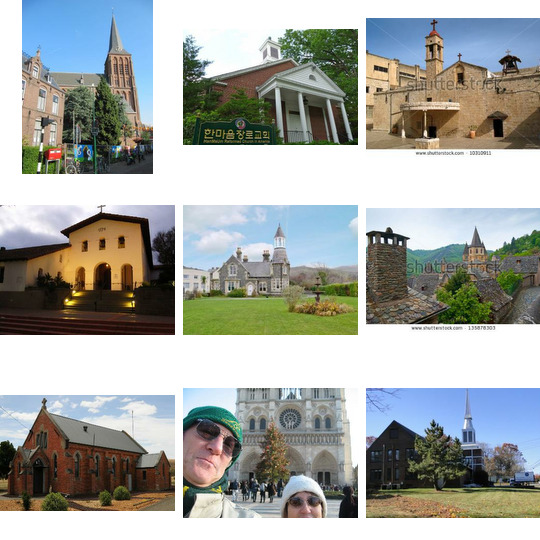}}
      & 
      \adjustbox{valign=c}{\includegraphics[height=\aspectheight]{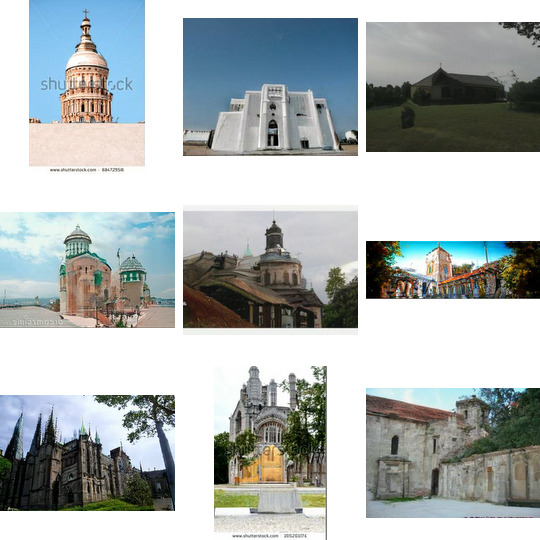}}
      &
      \adjustbox{valign=c}{\includegraphics[height=\squareheight]{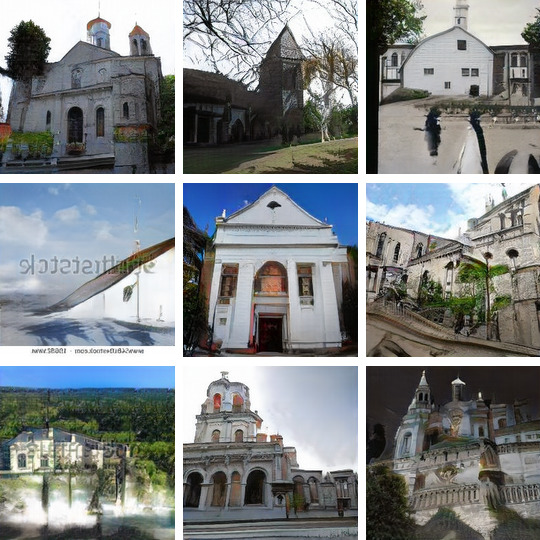}}
      &
      \adjustbox{valign=c}{\includegraphics[height=\squareheight]{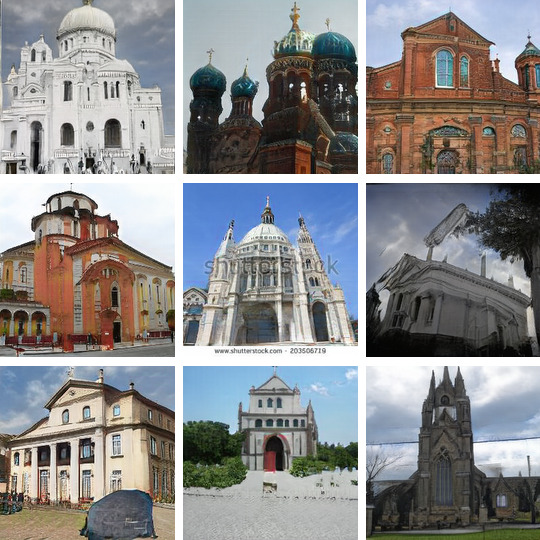}}
      \\  %\hline
      Data & DCTransformer & ProGAN & StyleGAN \\
      \end{tabular}
      \caption{Comparison between LSUN images and uncurated model samples for bedroom,tower and church-outdoor subsets. StyleGAN refers to StyleGAN1 for bedrooms, and StyleGAN2 for church-outdoor samples. DCTransformer produces variable aspect ratio samples with long-side resolution 384. BigGAN and VQ-VAE aretrained and sample at a fixed 256x256 resolution corresponding to a resized long-side crop of the input images. ProGAN and StyleGAN samples use truncation 1.0 to yield maximum diversity.}
      \label{fig:lsun_comparison}
      \end{center}
\end{figure*}

\end{document}